\documentclass[10pt]{article}

\usepackage[a4paper, margin=2cm]{geometry}
\usepackage[fleqn]{amsmath}
\usepackage{amssymb, amsfonts}
\usepackage{booktabs}

\usepackage{caption}
\usepackage[parfill]{parskip}

\usepackage{sectsty}
\sectionfont{\fontsize{11}{13}\selectfont}
\subsectionfont{\fontsize{11}{13}\selectfont}

\usepackage{breakcites}
\usepackage[breaklinks=true]{hyperref}  
\usepackage{amsthm}

\makeatletter
\newcommand*{\centerfloat}{%
  \parindent \z@
  \leftskip \z@ \@plus 1fil \@minus \textwidth
  \rightskip\leftskip 
  \parfillskip \z@skip}
\makeatother

\usepackage{xargs}                      
\usepackage[pdftex,dvipsnames]{xcolor} 

\usepackage{graphicx}
\usepackage[colorinlistoftodos,prependcaption,textsize=small]{todonotes}

\usepackage[T1]{fontenc} 
\usepackage[utf8]{inputenc}
\usepackage{lmodern}
\usepackage{import} 
\usepackage[normalem]{ulem}
\usepackage[ruled, vlined]{algorithm2e}

\newcommandx{\bart}[2][1=]{\todo[linecolor=Goldenrod,backgroundcolor=Goldenrod!25,bordercolor=Goldenrod,#1]{#2}}

\newcommandx{\rik}[2][1=]{\todo[linecolor=red,backgroundcolor=red!25,bordercolor=red,#1]{#2}}

\usepackage{tikz-cd}
\usetikzlibrary{shapes}

\newtheorem{remark}{Remark}[section]
\newtheorem{problem}{Problem}[section]

 \DeclareMathOperator{\tensor}{\otimes}
\DeclareMathOperator*{\argmin}{arg\,min}  \newcommand{\rr}{\mathbb{R}}
\newcommand{\OO}{O}
\graphicspath{{./imgs/}}

\newcommand{\inp}[2]{\left\langle#1,\,#2\right\rangle}

\begin{document}
\title{TTML: tensor trains for general supervised machine learning}
\author{Bart Vandereycken\thanks{University of Geneva. This work was supported by the SNSF under research project 192363.}\, and Rik Voorhaar\footnotemark[1]}
\date{}
\maketitle

\begin{abstract}
This work proposes a novel general-purpose estimator for supervised machine learning (ML) based on tensor trains (TT). The estimator uses TTs to parametrize discretized functions, which are then optimized using Riemannian gradient descent under the form of a tensor completion problem. Since this optimization is sensitive to initialization, it turns out that the use of other ML estimators for initialization is crucial. This results in a competitive, fast ML estimator with lower memory usage than many other ML estimators, like the ones used for the initialization.
\end{abstract}

\section{Introduction}

This paper proposes a new way to use tensor networks, in particular, tensor trains (TT) or matrix product states (MPS), as machine learning (ML) estimators for general supervised learning. While tensors networks have been applied in a variety of machine learning tasks, there has been much less attention on how to use them for general labeled datasets. Indeed, a popular area of applications of low-rank tensors is in deep learning where tensors can be used for compressing layers thereby lowering storage costs and increasing inference speed; see, e.g.,~\cite{tjandraCompressingRecurrentNeural2017,
novikovTensorizingNeuralNetworks2015}. They are also used in~\cite{daiTensorEmbeddingMethods2006a} to learn embeddings of very high-dimensional data so that traditional ML
techniques can then be applied to the lower-dimensional embeddings. Alternatively, high dimensionality can also be introduced on purpose when representing additional higher-order interactions between data as a tensor; see, e.g.,~\cite{novikovExponentialMachines2017,perrosPolyadicRegressionIts2017}. Finally, tensor completion in particular is useful
for recovering missing high-dimensional data \cite{liuTensorCompletionEstimating2013,kressnerLowrankTensorCompletion2014,hongGeneralizedCanonicalPolyadic2020}. All of these works have in common that tensors are used for tasks that are formulated from the start as a discrete tensor. In addition, the rationale in these applications for using low-rank tensors is in compressing high-dimensional data. For general datasets of smaller dimensionality, on the other hand, it is not clear how an ML estimator can be represented by a discrete tensor and what the benefit of low rank is. The main contribution of this paper is to close this gap.

Our method takes a different direction than the applications mentioned above. Instead of using low-rank tensors to model relatively simple relations on very high-dimensional
data, we use them to model complex non-linear patterns in low-dimensional data. In particular, we will learn a discretized version of a function $f\colon \rr^d\to\mathbb R$ on a suitable finite grid. The value of this function
on each grid cell can then be encoded by an order $d$ tensor. Both training and storing a dense tensor of this type is infeasible, and
we shall use low-rank tensor decompositions to make the problem computationally feasible. We propose to use TTs for this purpose, since they can be efficiently
trained from data using a Riemannian tensor completion algorithm. On a high level, this idea is similar to~\cite{kargasSupervisedLearningEnsemble2020,kargasNonlinearSystemIdentification2020} where CP tensors are used to learn functions $\rr^d\to\rr$ after discretization. However, their reported performance is worse than what we achieved on the same datasets, possibly due to the numerical difficulties in optimizing CP tensors.

We focus on two aspects: the discretization of the feature space, so that a discrete tensor can be used as estimator, and the learning of the estimator in a low-parametric family of tensors by local optimization. The motivation to use low-rank tensors for this task came from the identification of decision trees as sums of elementary tensors. We will show that, while decision trees naturally lead to simple low-rank tensors, one can actually learn better estimators within the same family that outperform the classical estimators with respect to model complexity and inference speed.

\subsection{Tensor completion as ML estimator}\label{sec:tensor completion as ML estimator}

As explained above, the main idea in this paper is to show that a low-rank tensor can be used to obtain a general purpose ML estimator that learns certain functions $f\colon \mathbb R^d\to \mathbb R$ from training data. To
see which functions $f\colon\mathbb R^d\to\mathbb R$ can be represented by a tensor, we first 
discretize the feature space $\mathbb R^d$. Such a discretization can be done using a set of \textit{thresholds} $\mathbf{t}_{(\alpha)} = (t_{(\alpha)}^1, \ldots, t_{(\alpha)}^{n_\alpha})$ for
each feature $1\leq \alpha\leq n$, with the convention
\begin{equation}\label{eq:convention ordered thresholds}
t_{(\alpha)}^1 \leq
t_{(\alpha)}^2 \leq \cdots \leq t_{(\alpha)}^{n_\alpha} = +\infty.
\end{equation}
 These
thresholds divide $\mathbb R$ into $n_{\alpha}+1$ bins of the form
$(t_{(\alpha)}^i,t_{(\alpha)}^{i+1}]$. The set of all these thresholds $\mathbf t=(\mathbf{t}_{(1)},\dots,\mathbf{t}_{(d)})$
thus divides $\mathbb R^d$ into a grid (this is described in some more detail in
Section~\ref{sec:dectree}). Then a tensor $\mathcal T\in
\rr^{|\mathbf{t}_{(1)}|}\tensor\cdots\tensor\rr^{|\mathbf{t}_{(d)}|}$ together with the thresholds
$\mathbf t$ defines a function $f_{\mathcal T,\mathbf t}\colon\mathbb R^d\to \mathbb R$ that is locally
constant on each grid cell. More precisely, $f_{\mathcal T,\mathbf t}$ is defined at $x \in \mathbb{R}^d$ as
\begin{equation}\label{eq:tensorfunc}
    f_{\mathcal T,\mathbf t}(x) = \mathcal T[j_1(x),\dots,j_d(x)],\qquad \text{where }\quad j_\alpha(x) = \max\{1\leq k\leq |\mathbf t_{(\alpha)}|\,\colon\, x[\alpha]\leq t_{(\alpha)}^k\}.
\end{equation}
Here, $x[\alpha]$ denotes the $\alpha$th entry of $x$. The notation for the $d$-dimensional entries of the tensor $\mathcal T$ is defined analogously. 

We thus consider a parametric family of functions $f_{\mathcal T,\mathbf t}$ that are constant on an irregular grid defined by thresholds $\mathbf{t}$. Supposing for the moment that $\mathbf t$ is given, we can learn the tensor $\mathcal T$, and hence
the function $f_{\mathcal T,\mathbf t}$, using training data $X=(x_1,\dots,x_N)\in\mathbb R^{d\times
N},\,y=(y_1,\dots,y_N) \in \mathbb R^N$ by minimizing an empirical loss function. This leads to the following \textit{tensor
completion problem}:

\begin{problem}\label{prob:tensorcompletion}
Let a set of thresholds $\mathbf t$, and training data $X,y$ be as above. We define the regression \textit{tensor completion problem} as minimizing the least-squares loss
\begin{equation}\label{eq:tensorcompletion-mse}
    \min_{\mathcal T\in \mathcal{M}}\sum_{i=1}^N \left(\mathcal T[j_1(x_i),\dots,j_d(x_i)] - y_i\right)^2,
\end{equation}
where $\mathcal{M}\subset \rr^{|\mathbf{t}_1|}\tensor\dots\tensor \rr^{|\mathbf{t}_d|}$ is a closed
subset of the space of tensors of (relatively) small dimension (that is, a `tensor format'). For
(binary) classification tasks where $y_i \in \{0,1\}$,  we instead consider the classification
tensor completion problem, which minimizes the cross-entropy loss
\begin{equation}
    \min_{\mathcal T\in \mathcal{M}}\sum_{i=1}^N -y_i\log\!\big(\sigma(\mathcal T[j_1(x_i),\dots,j_d(x_i)]) \big)
    -(1-y_i)\log\!\big(1-\sigma(\mathcal T[j_1(x_i),\dots,j_d(x_i)])\big),\label{eq:tensorcompletion-crossentropy}
\end{equation}
where $\sigma(t)=\exp(t)/(1+\exp(t))$ is the sigmoid function. These two loss functions correspond
respectively to the Gaussian and Bernoulli distributions in the context
of~\cite{hongGeneralizedCanonicalPolyadic2020}, and their work also states several more potential
loss functions that can be used in this context.
\end{problem}

\subsection{Main approach and outline of the paper}
Solving Problem~\ref{prob:tensorcompletion} in a good way is difficult in general. In this paper, we
present practical strategies to solve this problem so that we obtain useful and competitive ML estimators on real-world data. Our approach consists of four main `ingredients', which can more or less be chosen separately:
\begin{enumerate}
    \item A discretization of the feature space, or equivalently, a set of thresholds $\mathbf t$. 
    \item The choice of tensor format $\mathcal{M}\subset \rr^{|\mathbf{t}_1|}\tensor\cdots\tensor \rr^{|\mathbf{t}_d|}$.
    \item An algorithm to iteratively minimize the loss.
    \item A heuristic to initialize the tensor for the minimization algorithm.
\end{enumerate}

\paragraph{Feature space discretization.}  A good discretization of the feature space is essential when learning a discrete tensor as ML estimator. As an example and motivation, we first study decision trees in Section~\ref{sec:dectree}. They
naturally lead to a feature space discretization, since they can be losslessly represented as a tensor in low parametric form (more concretely, as a CP tensor). Choosing the discretization of the feature space is strictly speaking not part of the (classical)
tensor completion problem, but it is nevertheless essential to obtain good results in the setting of
prediction or classification using tensors. While the induced partitioning of decision trees can effectively be used as discretization of the feature space with tensor regression, we found that the most effective discretization of the
feature space is obtained by binning the data in roughly equally sized bins along each axis. We
discuss the choice of discretization  more detail in Section~\ref{sec:thresholds}. 

\paragraph{Tensor format.} There are many possibilities for the tensor subset $\mathcal{M}$ and a popular choice is a low-rank
format. For example, tensor networks that are based on a tree are well suited for optimization since
they can be locally optimized using alternating strategies or Riemannian optimization;
see~\cite{bachmayrTensorNetworksHierarchical2016} and  \cite{uschmajewGeometricMethodsLowRank2020}. Examples of
such networks are Tucker tensors, TTs, and hierarchical tensors.
In addition, the closure of such tensors of a certain rank is easily parametrized. In this paper, we
choose $\mathcal{M}$ as the set of TTs since it is a format that scales well to higher
dimensions and is relatively straightforward to implement. Finally, despite the representation of
a decision tree as a CP, the set of tensors of bounded CP rank is not a
closed set and its closure is not easily parametrized, which makes it more challenging for numerical treatment.

\paragraph{Optimization algorithm.}

To optimize the least-squares or cross-entropy loss in the TT format, we will use
Riemannian conjugate gradient descent (RCGD) as described in Section~\ref{sec:rcgd}. Other common
algorithms to optimize TTs are alternating least-squares (ALS) and DMRG, but in our
experiments they lead to very ill-conditioned subproblems when applied to real-world data. This is
because these two methods require a minimal amount of data samples for each tensor slice. The
datasets we use have relatively few data points that are not uniformly distributed, which puts severe restrictions on the size of the TT we can use with ALS and DMRG. This issue is
described in some more detail in Remark~\ref{rem:als-dmrg-problems}. The RCGD algorithm does not suffer from
this problem and is much more effective in solving Problem~\ref{prob:tensorcompletion}.

\paragraph{Initialization.}

While RCGD is very effective in finding good local minima for Problem~\ref{prob:tensorcompletion}, it needs to be properly initialized. Unless the rank of the TT
is very low, the tensor completion problem has many `bad' local minima. One way to deal with this
issue is to start solving the tensor completion problem with a very low rank TT, and then
adaptively increase the rank while resolving. While this gives decent results, we found that a much
more effective initialization is obtained by fitting the TT to match a pre-trained
estimator such as a random forest. We perform this fitting step using the TT-cross
algorithm~\cite{oseledetsTTcrossApproximationMultidimensional2010,savostyanovFastAdaptiveInterpolation2011}
as described in Section~\ref{sec:tt-cross}. Initialization can be performed with any kind of ML estimator. After further optimization of the loss, we can then match and sometimes beat the performance of
the chosen estimator when measured by inference speed or validation loss, while using less
parameters. This is shown empirically in Section~\ref{sec:experiments}

\subsection{Related work}

As mentioned in the introduction, learning ML estimators for general labeled datasets using low-rank tensors was also proposed in~\cite{kargasNonlinearSystemIdentification2020,kargasSupervisedLearningEnsemble2020}. The biggest difference with the current paper is that these works use CP tensors and more sophisticated ML techniques like bagging and boosting. However, the performance of the ML estimators based on CP is considerably worse than ours on TT. 
It is well known~\cite{hackbuschTensorSpacesNumerical2012} that optimizing over CP tensors is more challenging numerically: alternating minimization can suffer from long stagnation before convergence, known as swamping, and many robust optimization techniques, like cross approximations and Riemannian optimization, are not directly available in CP since the format is not based on nested matricizations. We therefore conjecture that using TT tensors (or general tensor networks based on nested matricizations) is crucial for obtaining good ML estimators.

When learning the ML estimators based on TT, we use several existing techniques that have become standard practice. In particular, after discretization we are left with a tensor completion problem over the set of bounded rank TT tensors. There exist many methods for this problem, mostly based on local optimization, hard thresholding or convex relaxations; see, e.g.~\cite{signorettoLearningTensorsFramework2014,grasedyckVariantsAlternatingLeast2015b,rauhutLowRankTensor2017a,grasedyckStableALSApproximation2019}. For reasons that we explain in Remark~\ref{rem:als-dmrg-problems}, we chose the Riemannian optimization method from~\cite{steinlechnerRiemannianOptimizationHighDimensional2016}. 

For initialization of the Riemannian optimization method, we use the TT-cross approximation method from~\cite{oseledetsTTcrossApproximationMultidimensional2010,savostyanovFastAdaptiveInterpolation2011,savostyanovQuasioptimalityMaximumvolumeCross2014a}. Recently, a parallelized version of this algorithm was developed in~\cite{dolgovParallelCrossInterpolation2020} which could improve the efficiency also for our setting. The idea of using TT-cross on a surrogate model so that all the fibers of the tensor are available was first proposed in~\cite{kapushevTensorCompletionGaussian2020}. There, the authors use a Gaussian process on a given incomplete tensor so that TT-cross can generate an initial guess for a tensor completion algorithm. We generalize this idea by fitting other ML estimators as surrogate model, like boosted trees, random forests or shallow neural networks; see~\cite{hastieElementsStatisticalLearning2009} for an overview of these classical ML estimators.

In contrast to most works (see our introduction and the overviews~\cite{cichockiTensorNetworksDimensionality2017a,songTensorCompletionAlgorithms2018}), the tensors in this paper only appear after a discretization of the feature space determined by the labelled data set. Discretizations are of course always needed when a discrete tensor approximates an infinite dimensional function. Tensor products of one dimensional discretizations, like Fourier series,  naturally give high-order tensors; see, e.g, ~\cite{wahlsLearningMultidimensionalFourier2014,bigoniSpectralTensorTrainDecomposition2016,imaizumiTensorDecompositionSmoothness2017}. Another possibility is tensorizations of very fine discretizations of low dimensional functions using finite elements; see, e.g.,~\cite{kazeevQuantizedTensorstructuredFinite2015}. Approximation rates of these continuous analogs of TT within certain function classes are available in~\cite{bigoniSpectralTensorTrainDecomposition2016,schneiderApproximationRatesHierarchical2014, kazeevQuantizedTensorstructuredFinite2017, griebelAnalysisTensorApproximation2021}. When learning an ML estimator from a relatively \emph{small} data set, however, these continuous tensor approximations are not suitable since the underlying functions are significantly less smooth. As example we give a random forest, which corresponds to a piecewise constant partitioning of the high-dimensional domain. This only allows for very simple discretizations, like quantile binning or k-means, as was one in~\cite{doughertySupervisedUnsupervisedDiscretization1995}. In case of image data, other possibilities for discretizations that lead to low-rank tensors are one-hot encoding or using fixed parametric families of functions, like wavelets; see~\cite{cohenExpressivePowerDeep2016,razinImplicitRegularizationTensor2021}. However, these also rely on the fact that images form a large dataset with a large degree of correlation between the features (pixels). This is not the case for our setting.

\section{Motivation: Decision trees as low-rank tensors}\label{sec:dectree}

We recall decision trees and show how they can be represented as a low-rank tensor. This serves as
motivation to consider more general ML models represented by low-rank tensors. We
refer to~\cite{hastieElementsStatisticalLearning2009} for a broader introduction into decision trees.

\begin{figure}[htb]
    \centering
    \def\svgwidth{0.9\textwidth}
    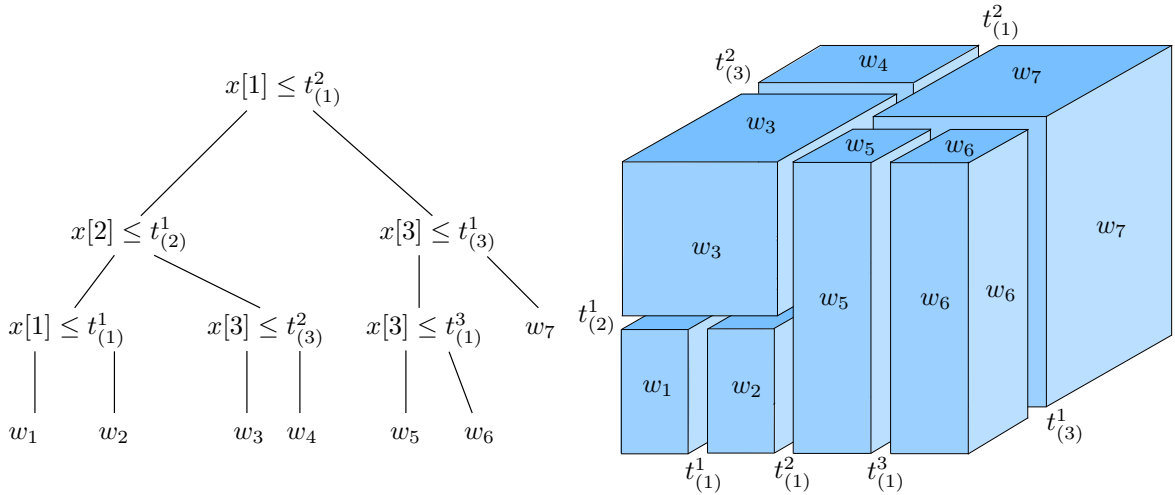
    \caption{Left: a decision tree $\rr^3\to\rr$. Right: the partitioning of $\rr^3$ determined by
        the tree on the left (where the outer corners represent points at $\pm\infty$). The
        function $h(x)$ is constant on each of the depicted rectangular regions $R_\ell$ where it
        takes on the indicated value $w_\ell$. For this tree, the thresholds are $\mathbf{t}_{(1)} =
        \{t_{(1)}^1,t_{(1)}^2,t_{(1)}^3,\infty \}$, $\mathbf{t}_{(2)} = \{t_{(2)}^1,\infty\}$,
        $\mathbf{t}_{(3)} = \{t_{(3)}^1,t_{(3)}^2,\infty\} $. }  
    \label{fig:dectree} 
\end{figure}

A (binary) decision tree $h\colon\rr^d\to \rr$ represents a recursive binary partitioning of the
feature space $\rr^d$ into labeled axis-aligned regions. In particular, at a non-leaf node of the
tree, we split $\rr^d$ into the two regions $\{ x \in \rr^d \colon x[\alpha] \leq t_{(\alpha)}^j \}$
and $\{x \in \rr^d \colon x[\alpha] > t_{(\alpha)}^j \}$ for some feature index $1\leq \alpha\leq d$
and threshold $t_{(\alpha)}^{j}$. Since a single feature $x[\alpha]$ can be used to split multiple
times, its associated thresholds are indexed by $j$. At each leaf $\ell$, we associate a label or
`decision' $w_\ell\in\rr$. To determine the value of $h(x)$ for some feature vector $x \in \rr^d$,
we start at the root of the tree and evaluate its corresponding inequality for $x$. If it is
satisfied, we proceed to the left child of the root, otherwise to the right child. This process is
repeated until we reach a leaf $\ell$ with its corresponding decision $w_\ell$. This determines the
value of $h(x)$. For an illustration, see Figure~\ref{fig:dectree}.

Each leaf $\ell$ corresponds to a region $R_\ell\subset \rr^d$ defined by the points satisfying the set of inequalities encountered
by traveling from the root to this leaf. Such a region is always an intersection of half-spaces and is therefore box-shaped, that is, all facets are normal to an axis. Note that these regions
are not necessarily bounded. The function associated to the decision tree is then given by
\begin{equation}\label{eq:decision tree as indicator function}
    h = \sum_{\ell\in L} w_\ell \, \mathbb{I}_{R_\ell},
\end{equation}
where $L$ is the set of leaves and $\mathbb{I}_{R_\ell}$ is the indicator function of the region $R_\ell$.

We now show that a decision tree $h\colon \mathbb R^d\to \mathbb R$ can be represented a low-rank tensor. 
Indeed, as is clear from~\eqref{eq:decision tree as indicator function} and the example in Figure~\ref{fig:dectree}, a decision tree is a function that is piecewise constant on a partitioning of $\rr^d$ into the boxes $R_\ell$. As explained in Section~\ref{sec:tensor completion as ML estimator}, we can thus represent $h$ with a tensor $\mathcal{T}$ and a set of thresholds $\mathbf{t}$. In fact, we can directly take over the thresholds from a decision tree, as long as we respect the ordering convention~\eqref{eq:convention ordered thresholds}. Up to a trivial reordering, this can of course always be done. We have thus defined the set of all thresholds  $\mathbf{t}_{(\alpha)} = (t_{(\alpha)}^1, \ldots, t_{(\alpha)}^{n_\alpha})$ for all features $1 \leq \alpha \leq d$.

It remains to construct the tensor $\mathcal T\in \rr^{|\mathbf{t}_{(1)}|}\tensor\cdots\tensor\rr^{|\mathbf{t}_{(d)}|}$ so that the function $f_{\mathcal T,\mathbf t}\colon \rr^d\to \rr$ defined in~\eqref{eq:tensorfunc} coincides with the decision tree $h$, that is, $f_{\mathcal T,\mathbf t} = h$. It is not hard to see that this means that the elements of $\mathcal T$ have to be the decision values $w_\ell$. Note that many entries of
$\mathcal T$ will be the same, since each region $R_\ell$ is typically divided into multiple grid
cells. In the example of Figure~\ref{fig:dectree}, we have that $\mathcal T[3,1,3]$ corresponds to
the region $(t_{(1)}^2,t_{(1)}^3]\times (-\infty,t_{(2)}^1]\times (t_{(3)}^2,\infty)$ where
$f_{\mathcal T,\mathbf t}$ takes value $w_7$. Leaf 7 corresponds to the region
$R_7=(t_{(1)}^2,\infty)\times \rr\times (t_{(3)}^1,\infty)$ and thus $\mathcal T[4,1,3]$,
$\mathcal T[3,2,3]$, and $\mathcal T[3,1,2]$ have, for example, the same value $w_7$. The complete tensor in Figure~\ref{fig:dectree} is given by its two slices as follows:

\[
\mathcal T[:,1,:] = \begin{pmatrix}
    w_1 & w_1 & w_1 \\
    w_2 & w_2 & w_2 \\
    w_5 & w_7 & w_7 \\
    w_6 & w_7 & w_7
\end{pmatrix},
\qquad 
\mathcal T[:,2,:] = \begin{pmatrix}
    w_3 & w_3 & w_4 \\ 
    w_3 & w_3 & w_4 \\
    w_5 & w_7 & w_7 \\
    w_6 & w_7 & w_7
\end{pmatrix}.
\]

Having obtained a tensor $\mathcal{T}$ from $h$, we now show that $\mathcal{T}$ is of low rank. We will do this by writing $\mathcal T$ as a sum of at most $|L|$ elementary tensors that are of rank one, where $L$ is the set of leaves in the decision
tree. These elementary tensors are obtained in terms of the \textit{active thresholds} at each leaf. Each leaf $\ell\in L$
corresponds to a region
\[
    R_{\ell} = (a_1,b_1]\times\dots\times (a_d,b_d],
\]
where possibly $a_\alpha=-\infty$ and $b_\alpha=+\infty$. We define a threshold $t_{(\alpha)}^j$ to be \textit{active at the leaf
    $\ell$} if $(t_{(\alpha)}^j,t_{(\alpha)}^{j+1}]\subset (a_\alpha,b_\alpha]$. For convenience, we formally set $t_{(\alpha)}^0=-\infty$. For example, in
the decision tree of Figure~\ref{fig:dectree},  the active thresholds for leaf $7$ are
$\{t_{(1)}^2,t_{(1)}^3,t_{(2)}^0,t_{(2)}^1,t_{(3)}^1,t_{(3)}^2\}$. For each feature $\alpha$ and leaf $\ell\in L$, we then define the vector
$v_\ell^{(\alpha)} \in \rr^{|\mathbf t_{(\alpha)}|}$ entry-wise as
\[
    v_\ell^{(\alpha)} [j] = \left\{ \begin{array}{ll}
        1, & \text{if $t_{(\alpha)}^j$ is active at the leaf $\ell$;} \\
        0, & \text{otherwise.}
    \end{array}\right.
\]
For our example, we obtain $v_7^{(1)} = (0,0,1,1)$, $v_7^{(2)}=(1,1)$, and $v_7^{(3)}=(0,1,1)$. Consider now the tensor
\[
    \mathcal T_\ell = w_\ell \cdot v_\ell^{(1)}\tensor \dots\tensor v_\ell^{(d)},
\]
which is clearly of rank at most one.  
It is readily verified using~\eqref{eq:tensorfunc} that then $f_{\mathcal T_\ell,\mathbf t}$ is supported on $R_\ell$ where it has constant value $w_\ell$. Hence, from~\eqref{eq:decision tree as indicator function} we see that
\begin{equation}\label{eq:dectree-CP}
    \mathcal T = \sum_{\ell\in L}w_\ell \cdot v_\ell^{(1)}\tensor \dots\tensor v_\ell^{(d)}
\end{equation}
allows us to represent the decision tree as $h=f_{\mathcal T, \mathbf{t}}$.

\begin{figure}[htb]
    \centering
    \def\svgwidth{0.8\textwidth}
    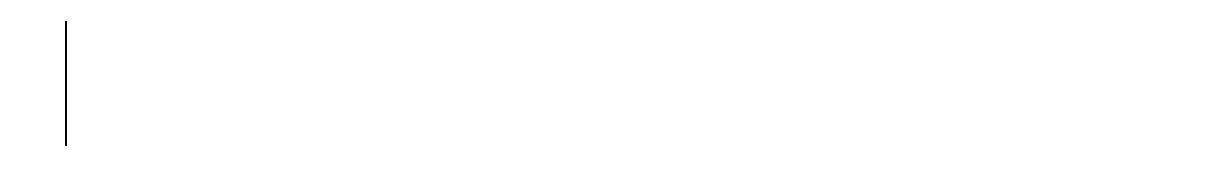
    \caption{A linear decision tree $\rr^2\to\rr$. }
    \label{fig:dectree_counterexample}
\end{figure}

The CP rank of $\mathcal T$ is the minimal number of summands of non-zero elementary tensors needed to express $\mathcal T$. The formula
above shows that the CP rank  of $\mathcal T$ is at most $|L|$, but it is in fact lower if the tree is not linear. This is because
if two leaves $\ell,\ell'$ share the same parent, then $\mathcal T_\ell +\mathcal T_{\ell'}$ is of rank 1. Indeed, if the parent of $\ell$
and $\ell'$ were split for feature $\alpha$, then we have that $v_\ell^{(\nu)} = v_{\ell'}^{(\nu)}$ for all $\nu\neq \alpha$ and thus
\[
    \mathcal T_\ell +\mathcal T_{\ell'} = v_\ell^{(1)}\tensor \dots\tensor v^{(\alpha-1)}_\ell \tensor \left(w_\ell v_\ell^{(\alpha)} + w_{\ell'}v^{(\alpha)}_{\ell'}\right)\tensor v^{(\alpha+1)}_\ell\tensor\dots\tensor v^{(d)}_\ell.
\]
The CP rank of $\mathcal T$ is therefore also bounded by \textit{the number of nodes that have a leaf attached} rather than just the
number of leaves. In general, this is just an upper bound. For example, the tensor associated to the decision tree shown in
Figure~\ref{fig:dectree_counterexample} is of rank 2 since it satisfies
\[
    \mathcal T = w_1\left(\sum_{i=1}^n e_i\right)\tensor e_1 + \left(\sum_{i=1}^n w_{i+1}e_i\right)\tensor e_2,
\]
where $e_i$ is the $i$th standard basis vector.

\begin{remark}\label{rem:cp_rank}
    Even though the CP rank has a useful upper bound, there is no similar upper bound on the Tucker
    rank. The Tucker rank $(r_1,\dots,r_d)$ of $\mathcal T$ is a tuple encoding the minimal
    dimensions of spaces $V_{\alpha} \subset \rr^{|\mathbf t_{(\alpha)}|}$ such that $\mathcal
    T\subset V_{(1)}\tensor\dots\tensor V_{(d)}$. Depending on the tree structure, we can in general
    have that the Tucker rank is maximal: $(r_1,\dots,r_n)=(|\mathbf t_{(1)}|,\dots,|\mathbf
    t_{(d)}|)$. Furthermore, the singular values are on the same order as the weights $w_i$, and
    hence do not decay fast. This means that these tensors cannot be well-approximated (or compressed) using a
    HOSVD decomposition. 
\end{remark}

From Remark~\ref{rem:cp_rank} above we conclude that compressing the tensor associated to a decision
tree in a tensor format like CP or Tucker is probably not useful. For example, as we will also
note in Remark~\ref{rem:rf_tt_compress}, we cannot efficiently express most decision trees in the
TT format either. Furthermore, in ML estimation tasks, we care about minimizing the test error and in principle not about compression
error. However, we will see that the underlying tensor from a decision tree can be useful as initialization when solving Problem~\ref{prob:tensorcompletion}; see Section~\ref{sec:tt-cross}.

Finally, we remark that we can also see random forests as tensors in the CP format. Recall that in a
random forest we independently train a number of decision trees, each on a different subset of the
training data. We then average out the predictions of all the decision trees to obtain a useful
estimator. This is therefore a weighted sum of decision trees, each of which we can represent by a
CP tensor. Similarly in a boosted forest, we train decision trees sequentially and combine them with
a weighted sum as well. It is important to note however that the set of thresholds $\mathbf t$ is
different for each decision tree. To represent the random/boosted forest as a CP tensor we have to
merge the sets of thresholds of all the constituent decision trees. The result is that both the CP
rank and the number of thresholds increase linearly with the number of trees in the forest,
resulting in a quadratic increase in storage costs.

\section{The tensor train decomposition}\label{sec:tt}

The tensor train (TT) decomposition is a popular low-rank decomposition of tensors. We recall
basic properties of this decomposition and describe a solution to the tensor completion problem for
this format. Most of the content in this section is fairly standard; see,
e.g.,~\cite{hackbuschTensorSpacesNumerical2012,oseledetsTensorTrainDecomposition2011}. For the
tensor completion problem and Riemannian structure of the TT format we mainly
follow~\cite{steinlechnerRiemannianOptimizationHighDimensional2016}.

A TT decomposition of a tensor $\mathcal T\in \rr^{n_1}\tensor\dots\tensor \rr^{n_d}$ 
consists of $d$ order 3 \textit{core tensors} $\mathcal C_{(\alpha)}$ of the form
\[
    \mathcal C_{(1)}\in\rr^{1\times n_1\times r_1},\,
    \mathcal C_{(2)}\in\rr^{r_1\times n_2\times r_2},\,\dots,\,
    \mathcal C_{(d-1)}\in\rr^{r_{d-2}\times n_{d-1}\times r_{d-1}},\,
    \mathcal C_{(d)}\in\rr^{r_{d-1}\times n_d\times 1}.
\]
We denote the slice $C_{(\alpha)}[i] := C_{(\alpha)}[\colon,\, i,\colon]$, which is for each $i$ a matrix of
    size $r_{\alpha-1}\times r_{\alpha}$. The tensor $\mathcal T$ is then decomposed element-wise as
\begin{equation}\label{eq:TT-def}
    \mathcal T[i_1,\dots,i_d] = \mathcal C_{(1)}[i_1] \, \mathcal C_{(2)}[i_2] \, \cdots \, \mathcal C_{(d-1)}[i_{d-1}] \, \mathcal C_{(d)}[i_d].
\end{equation}
This decomposition is shown in Penrose graphical notation below. The labels on the edges indicate
dimensions, and the two one-dimensional edges are omitted.
\[
    \begin{tikzcd}[column sep=3em, row sep=4ex]
        |[draw=black, rectangle]| \mathcal C_{(1)} \ar[r, dash, "r_1"] \ar[d, dash, "n_1"] &
        |[draw=black, rectangle]| \mathcal C_{(2)} \ar[r, dash, "r_2"] \ar[d, dash, "n_2"]& \cdots &
        |[draw=black, rectangle]| \mathcal C_{(d-1)} \ar[l, dash, "r_{d-2}"'] \ar[r, dash, "r_{d-1}"] \ar[d, dash, "n_{d-1}"] &
        |[draw=black, rectangle]| \mathcal C_{(d)} \ar[d, dash, "n_d"]\\
        {}&{}&{}&{}&{}
    \end{tikzcd}
\]
The tuple $\mathbf r=(r_1,\dots,r_{d-1})$ is called the \textit{TT-rank}. The set of all tensors
that can be written as a TT of \textit{exactly} rank $\mathbf r$ form a (non-closed) embedded submanifold
$\mathcal M_{= \mathbf r}\subset \rr^{n_1}\tensor\dots\tensor\rr^{n_d}$. In Section~\ref{sec:man_struct} we
will study the Riemannian geometry of this manifold. The closure of $\mathcal M_{= \mathbf r}$ consists of all TT of rank \textit{at most} rank $\mathbf r$, which we denote by $\mathcal M_{\leq \mathbf r}$. Each TT in $\mathcal M_{\leq \mathbf r}$ can be written in the form~\eqref{eq:TT-def}. We note that $\mathcal M_{\leq \mathbf r}$ is not a submanifold.

\subsection{Orthogonalization and rank truncation}\label{sec:ortho and hosvd}

An important property of the TT decomposition is that we can \textit{orthogonalize} the cores. Each core $\mathcal C_{(\alpha)}$
has a left matricization $\mathcal C_{(\alpha)}^L\in \rr^{r_{\alpha-1}n_\alpha\times r_\alpha}$ and a right matricization
$\mathcal C_{(\alpha)}^R\in \rr^{r_{\alpha-1}\times n_\alpha r_\alpha}$ by flattening or `matricizing' the order 3 tensor $\mathcal C_{(\alpha)}$ to a matrix. The core is left-orthogonal if the columns of $C_{(\alpha)}^L$ are orthonormal: ${(C_{(\alpha)}^L)^\top C_{(\alpha)}^L = I_{r_\alpha}}$. Similarly, right-orthogonal means $C_{(\alpha)}^R
    {(C_{(\alpha)}^R)^\top = I_{r_{\alpha-1}}}$. We call a TT \textit{orthogonalized at mode $\mu$} if all cores $C_{(\alpha)}$ are
left-orthogonal for $\alpha<\mu$ and right-orthogonal for $\alpha>\mu$. A TT is left-orthogonalized if it is orthogonalized at
mode $d$, similarly it is right-orthogonalized if it is orthogonalized at mode $1$.

\begin{figure}[htb]
\[
    \hspace{6.5em} \begin{tikzcd}[remember picture, column sep=4em, row sep=6ex, cells={nodes={draw=black,rectangle}}]
        \mathcal C_{(1)} \ar[r, dash, "r_1"] \ar[d, dash, "n_1"] &
        \mathcal C_{(2)} \ar[r, dash, "r_2"] \ar[d, dash, "n_2"] &
        \mathcal C_{(3)} \ar[r, dash, "r_3"] \ar[d, dash, "n_3"] &
        \mathcal C_{(4)} \ar[r, dash, "r_4"] \ar[d, dash, "n_4"] &
        \mathcal C_{(5)} \ar[d, dash, "n_5"] \\
        \mathcal C_{(1)} \ar[r, dash, "r_1"] &
        \mathcal C_{(2)} \ar[r, dash, "r_2"] &
        \mathcal C_{(3)} \ar[r, dash, "r_3"] &
        \mathcal C_{(4)} \ar[r, dash, "r_4"] &
        \mathcal C_{(5)}
    \end{tikzcd}
    \tikz[remember picture, overlay]{
        \fill[blue,opacity=0.2] ([xshift=-5,yshift=5]\tikzcdmatrixname-1-1.north west) rectangle ([xshift=5,yshift=-5]\tikzcdmatrixname-1-2.south east);
        \node[] at ([xshift=-20,yshift=12]\tikzcdmatrixname-1-2.north west) {\color{blue}$\mathcal C_{\leq 2}$};
        \fill[blue,opacity=0.2] ([xshift=-5,yshift=5]\tikzcdmatrixname-1-4.north west) rectangle ([xshift=5,yshift=-5]\tikzcdmatrixname-1-5.south east);
        \node[] at ([xshift=-20,yshift=12]\tikzcdmatrixname-1-5.north west) {\color{blue}$\mathcal C_{\geq 4}$};
        \fill[red,opacity=0.2] ([xshift=-5,yshift=5]\tikzcdmatrixname-1-1.north west) rectangle ([xshift=5,yshift=-5]\tikzcdmatrixname-2-1.south east);
        \node[] at ([xshift=-25,yshift=-15]\tikzcdmatrixname-1-1.south west) {\color{red}$\mathcal C_{(1)}^\top\mathcal C_{(1)}$};
    }
\]
\[
    \hspace{4.5em} =\begin{tikzcd}[column sep=4em, row sep=6ex, cells={nodes={draw=black,rectangle}}]
        \mathcal C_{(2)} \ar[r, dash, "r_2"] \ar[d, dash, "n_2"] \ar[d, start anchor=west, end anchor=west,dash, bend right=50, "r_1"'] &
        \mathcal C_{(3)} \ar[r, dash, "r_3"] \ar[d, dash, "n_3"] &
        \mathcal C_{(4)} \ar[d, dash, "n_4"'] \ar[d, dash, start anchor=east, end anchor=east, bend left=50, "r_5"]  \\
        \mathcal C_{(2)} \ar[r, dash, "r_2"] &
        \mathcal C_{(3)} \ar[r, dash, "r_4"] &
        \mathcal C_{(4)}
    \end{tikzcd}
    \quad =\quad
    \begin{tikzcd}[column sep=4em, row sep=6ex, cells={nodes={draw=black,rectangle}}]
        \mathcal C_{(3)} \ar[d, dash, "r_2"', start anchor=west, end anchor=west, bend right=50] \ar[d, dash,"n_3" description] \ar[d, dash, "r_3", start anchor=east, end anchor=east,  bend left=50]\\
        \mathcal C_{(3)}
    \end{tikzcd}
\]
\caption{The inner product $\langle \mathcal{T}, \mathcal{T} \rangle$ of a TT tensor $\mathcal{T}$. Since $\mathcal{T}$ is orthogonalized at mode $3$, the contractions of several cores can be simplified to identity (indicated in red for the first core) so that the final result equals $\langle \mathcal{C}_{(3)}, \mathcal{C}_{(3)} \rangle$.}
\label{fig:inner product TT}
\end{figure}
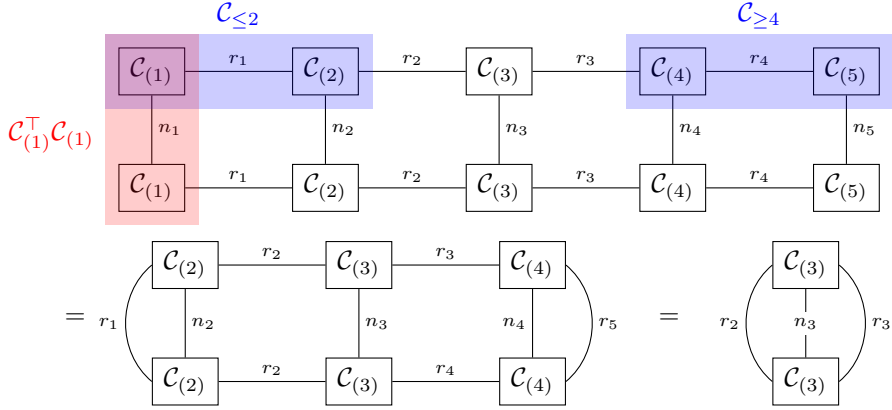

If a TT is orthogonalized at mode $\mu$, then we also have that $\mathcal C_{\leq \alpha}^\top
\mathcal C_{\leq \alpha}=I_{r_\alpha}$, where $\mathcal C_{\leq \alpha}$ is obtained by contracting
all cores $\mathcal C_{(\beta)}$ with $\beta\leq \alpha$. This is because on the very left we have
$\mathcal C_{(1)}$ which is already a matrix, and thus $\mathcal C_{(1)}^\top\mathcal
C_{(1)}=I_{r_1}$. Similarly for $\alpha>\mu$, we have  $\mathcal C_{\geq \alpha} \mathcal C_{\geq
\alpha}^\top=I_{r_{\alpha-1}}$. This is shown below for $d=5$ and a TT orthogonalized at mode
$\mu=3$. This also shows that if a TT tensor $\mathcal T$ is orthogonalized at mode $\mu$, then we can
compute its Euclidean norm as $\|\mathcal T\|=\|\mathcal C_{(\mu)}\|$. See Figure~\ref{fig:inner product TT} for a graphical illustration of this idea.

Any TT can easily be orthogonalized through a QR-procedure. To left-orthogonalize, for example, we perform the following procedure to each core from left
to right: Compute the QR decomposition of the left matricization $\mathcal C_{(\alpha)}^L=Q R$. Set
$\mathcal C_{(\alpha)}$ to the tensorisation $\mathcal Q_{(\alpha)}$ of $Q$. Multiply the matrix $R$ into the tensor $\mathcal
    C_{(\alpha+1)}$. These steps are also shown below.
\[
    \newcommand{\toWidth}{\hspace{-0.35em}}
    \begin{tikzcd}[column sep=2em, row sep=4ex, cells={nodes={draw=black}}]
        |[draw=white]| &
        \mathcal C_{(\alpha)} \ar[l,dash,"r_{\alpha-1}"'] \ar[r,dash,"r_{\alpha}"] \ar[d,dash,"n_\alpha"] &
        \mathcal C_{(\alpha+1)} \ar[d,dash,"n_{\alpha+1}"] \ar[r,dash,"r_{\alpha+1}"]&
        |[draw=white]|\\|[draw=white]|&|[draw=white]|&|[draw=white]|
    \end{tikzcd}\toWidth=\toWidth
    \begin{tikzcd}[remember picture, column sep=2em, row sep=4ex, cells={nodes={draw=black}}]
        |[draw=white]| &
        \mathcal Q_{(\alpha)} \ar[l,dash] \ar[r,dash,"r_{\alpha}"] \ar[d,dash] &
        R \vphantom{C_{\alpha}}\ar[r,dash,"r_{\alpha}"]&
        \mathcal C_{(\alpha+1)} \ar[d,dash] \ar[r,dash]&
        |[draw=white]|\\|[draw=white]|&|[draw=white]|&|[draw=white]|&|[draw=white]|
    \end{tikzcd}
    \tikz[remember picture, overlay]{
        \draw[black,dashed] ([xshift=-3,yshift=5]\tikzcdmatrixname-1-3.north west) rectangle ([xshift=3,yshift=-4]\tikzcdmatrixname-1-4.south east);
    }
    \toWidth=\toWidth
    \begin{tikzcd}[column sep=2em, row sep=4ex, cells={nodes={draw=black}}]
        |[draw=white]| &
        \mathcal Q_{(\alpha)} \ar[l,dash,"r_{\alpha-1}"'] \ar[r,dash,"r_{\alpha}"] \ar[d,dash,"n_\alpha"] &
        \mathcal C'_{(\alpha+1)} \ar[d,dash,"n_{\alpha+1}"] \ar[r,dash,"r_{\alpha+1}"]&
        |[draw=white]|\\|[draw=white]|&|[draw=white]|&|[draw=white]|
    \end{tikzcd}
\] If we want to instead right-orthogonalize a TT, then we perform the orthogonalization right-to-left, and each
individual step in the orthogonalization is transposed with respect to the picture above. Using that a QR factorization of an $m\times n$ matrix costs $mn^2$ flops, the cost of a complete left-to-right or right-to-left sweep is given by
\begin{equation}\label{eq:orth-complexity}
    n_1r_1^2+n_dr_{d-1}^2 + \sum_{\alpha=2}^{d-1}n_\alpha r_{\alpha-1}r_\alpha (r_{\alpha-1}+r_\alpha ).
\end{equation}
To orthogonalize at mode $\mu$ we
left-orthogonalize all the cores $\mathcal C_{(\alpha)}$ with $\alpha<\mu$ and right-orthogonalize those with $\alpha>\mu$.

Instead of a QR decomposition, one can also use the singular value decomposition (SVD) above. While SVD is slightly more expensive than QR, it allows us to reduce the ranks $\mathbf{r}$ of a given TT in a quasi-optimal manner. For example, in the diagram below, if $\mathcal C_{(\alpha)}^L = U\Sigma V$ is
the SVD and $U'\Sigma'V'$ the rank-$r'_{\alpha}$ truncated SVD, then we update $\mathcal
C_{(\alpha)}$ by the tensorisation $\mathcal U_{\alpha}$ of $U$, and we absorb $\Sigma'V'$ into the next core:
\[\hspace{-0.6em}
    \newcommand{\toWidth}{\hspace{-0.6em}}
    \begin{tikzcd}[column sep=2em, row sep=4ex, cells={nodes={draw=black}}]
        |[draw=white]| &
        \mathcal C_{(\alpha)} \ar[l,dash,"r_{\alpha-1}'"'] \ar[r,dash,"r_{\alpha}"] \ar[d,dash,"n_\alpha"] &
        \mathcal C_{(\alpha+1)} \ar[d,dash,"n_{\alpha+1}"] \ar[r,dash,"r_{\alpha+1}"]&
        |[draw=white]|\\|[draw=white]|&|[draw=white]|&|[draw=white]|
    \end{tikzcd}\toWidth\to\toWidth
    \begin{tikzcd}[remember picture, column sep=2em, row sep=4ex, cells={nodes={draw=black}}]
        |[draw=white]| &
        \mathcal U_{(\alpha)} \ar[l,dash] \ar[r,dash,"r_{\alpha}'"] \ar[d,dash] &
        \Sigma'V' \vphantom{C_{\alpha}}\ar[r,dash,"r_{\alpha}"]&
        \mathcal C_{(\alpha+1)} \ar[d,dash] \ar[r,dash]&
        |[draw=white]|\\|[draw=white]|&|[draw=white]|&|[draw=white]|&|[draw=white]|
    \end{tikzcd}
    \tikz[remember picture, overlay]{
        \draw[black,dashed] ([xshift=-3,yshift=5]\tikzcdmatrixname-1-3.north west) rectangle ([xshift=3,yshift=-5]\tikzcdmatrixname-1-4.south east);
    }\toWidth=\toWidth
    \begin{tikzcd}[column sep=2em, row sep=4ex, cells={nodes={draw=black}}]
        |[draw=white]| &
        \mathcal U_{(\alpha)} \ar[l,dash,"r_{\alpha-1}'"'] \ar[r,dash,"r_{\alpha}'"] \ar[d,dash,"n_\alpha"] &
        \mathcal C'_{(\alpha+1)} \ar[d,dash,"n_{\alpha+1}"] \ar[r,dash,"r_{\alpha+1}"]&
        |[draw=white]|\\|[draw=white]|&|[draw=white]|&|[draw=white]|
    \end{tikzcd}
\] 
Doing this in a left-to-right fashion to all cores both left-orthogonalizes the TT and reduces its TT-rank from $\mathbf r$ to $\mathbf r'$. We will denote this truncation by the map $\mathrm{HOSVD}(\mathbf r,\mathbf r')\colon\mathcal
M_{\leq \mathbf r}\to \mathcal M_{\leq \mathbf r'}$ which is quasi-optimal in the sense that 
\begin{equation}\label{eq:quasi-optimality}
    \|\mathcal T - \mathrm{HOSVD}(\mathbf r,\mathbf r')(\mathcal T) \| \leq \sqrt{d-1}\|\mathcal T-\mathcal P_{\mathcal M_{\leq \mathbf r'}}(\mathcal T)\|,
\end{equation}
where $\mathcal P_{\mathcal M}(\mathcal T)$ denotes the best approximation of $\mathcal T$ in the closed subset $\mathcal M$; see
\cite[\S 12.2.7]{hackbuschTensorSpacesNumerical2012}. Assuming the SVD of an $m\times n$ matrix costs $2mn^2+2n^3$ flops, computing $\mathrm{HOSVD}$ costs
\begin{equation}\label{eq:hosvd-cost}
    2n_1r_1^2+2r_1^3+\left(\sum_{\alpha=2}^{d-1}r_{\alpha-1}'r_{\alpha-1}n_\alpha r_\alpha +2r_{\alpha-1}'n_\alpha r_\alpha^2+2r_\alpha^3\right) + r'_{d-1}r_{d-1}n_d.
\end{equation}

\begin{remark}\label{rem:rf_tt_compress}
    A tensor in CP format can be written easily in TT format; see
\cite[\S 8.5.2]{hackbuschTensorSpacesNumerical2012}. In particular, let 
    \[
        \mathcal T = \sum_{\ell\in L}w_\ell \cdot v_\ell^{(1)}\tensor \dots\tensor v_\ell^{(d)},
    \]    
    as in~\eqref{eq:dectree-CP} be a CP tensor that represents a decision tree, 
    then we can define the cores of the corresponding (left-orthogonal) TT as
    \[
        \mathcal C_{(\alpha)}[i,\ell,j] = \delta_{ij}v_\ell^{(\alpha)}[i],\qquad \mathcal C_{(1)}[1,\ell,i] = v_{\ell}^{(1)}[i],\qquad \mathcal C_{(d)}[i,\ell,1] = v_\ell^{(d)}[i]\cdot w_\ell.
    \]
    This is unfortunately not very useful in practice. The TT rank is equal to $(r,\dots,r)$
    with $r$ the CP rank of $\mathcal T$, which typically scales linearly with the size of the
    decision tree, as discussed in Section~\ref{sec:dectree}. Furthermore, the HOSVD gives in
    practice very bad approximations, since most singular values are $O(1)$. This is
    because the vectors $v_\ell^{(\alpha)}$ have norm $O(n_\alpha)$ and are close to orthogonal.
    Any reasonable truncation of $\mathcal{T}$ by HOSVD will thus have a very high TT-rank. Nevertheless, and quite surprisingly, we can still get useful
    approximations of random forests (sums of decision trees) with small TT-rank using the TT-cross algorithm. This is discussed in
    Section~\ref{sec:tt-cross}.
\end{remark}

\subsection{Manifold structure}\label{sec:man_struct}

To perform Riemannian optimization on a manifold $ M$, we need three main ingredients: a description
of the tangent bundle $TM$ and its metric, a retraction $R\colon TM\to M$, and a vector transport
map $TM\times_M TM\to TM$; see, e.g., the general overview~\cite{absilOptimizationAlgorithmsMatrix2008}. For the manifold of TTs, the usual choice for retraction and vector transport are the HOSVD truncation and the orthogonal projection onto the tangent space. We will briefly recall the description of these `ingredients' as introduced
by~\cite{steinlechnerRiemannianOptimizationHighDimensional2016}. 

\paragraph{Tangent vectors.} Let $(\mathcal
U_{(1)},\dots,\mathcal U_{(d)})$ be the left-orthogonal cores of a TT tensor $\mathcal T\in\mathcal
M_{=\mathbf r}$. Similarly, $(\mathcal V_{(1)},\dots,\mathcal V_{(d)})$ are its right-orthogonal cores. Then tangent vectors $\delta\mathcal{Y} \in T_\mathcal
T\mathcal M_{= \mathbf r}\subset \rr^{n_1}\tensor\cdots\tensor \rr^{n_d}$ are given by tensors of the form
\begin{align}
    \begin{split}
        \delta\mathcal Y[i_1,\dots,i_d] = &\quad \delta \mathcal Y_{(1)}[i_1] \, \mathcal V_{(2)}[i_2]\, \cdots \, \mathcal V_{(d)}[i_d]\\
        +&\quad \mathcal U_{(1)}[i_1]\, \delta \mathcal Y_{(2)}[i_2]\, \mathcal V_{(3)}[i_3]\, \cdots \, \mathcal V_{(d)}[i_d]\\
        +\dots+&\quad \mathcal U_{(1)}[i_1]\, \mathcal U_{(2)}[i_2]\, \cdots \, \mathcal U_{(d-2)}[i_{d-2}]\, \delta \mathcal Y_{(d-1)}[i_{d-1}] \, \mathcal V_{(d)}[i_d]\\
        +&\quad \mathcal U_{(1)}[i_1]\, \mathcal U_{(2)}[i_2]\, \cdots \, \mathcal U_{(d-1)}[i_{d-1}]\, \delta \mathcal Y_{(d)}[i_{d}]
    \end{split}\label{eq:tangent-form}
\end{align}
where $\delta \mathcal Y_{(\alpha)}$ has the
same shape as $\mathcal U_{(\alpha)}$. We call the tensors $\delta Y_{(\alpha)}$ \textit{first-order
variations}. If they satisfy the \textit{gauge condition} $(\mathcal U_{(\alpha)}^L)^\top
\delta \mathcal Y^L_{(\alpha)} = 0$ for all $\alpha<d$, then they are uniquely defined from~\eqref{eq:tangent-form} for each tangent vector $\delta\mathcal{Y}$.

Note that the representation~\eqref{eq:tangent-form} of $\delta \mathcal Y$ is itself a rank-$2\mathbf r$
TT: thanks to the gauge condition, we can indeed write
\begin{equation}\label{eq:tangent vector as rank 2r TT}
    \mathcal Y[i_1,\dots,i_d] = (\delta \mathcal Y_{(1)}[i_1]\, \mathcal U_{(1)}[i_1])
    \begin{pmatrix}
        \mathcal V_{(2)}[i_2]        & 0                     \\
        \delta \mathcal Y_{(2)}[i_2] & \mathcal U_{(2)}[i_2]
    \end{pmatrix}\cdots
    \begin{pmatrix}
        \mathcal V_{({d-1})}[i_{d-1}]        & 0                             \\
        \delta \mathcal Y_{({d-1})}[i_{d-1}] & \mathcal U_{({d-1})}[i_{d-1}]
    \end{pmatrix}
    \begin{pmatrix}
        \mathcal V_{({d})}[i_{d}] \\\delta\mathcal Y_{({d})}[i_{d}]
    \end{pmatrix}.
\end{equation}

\paragraph{Metric.} As metric for $\mathcal M_{= \mathbf r}$, we restrict the Euclidean metric on $\rr^d$ to $T\mathcal M_{= \mathbf r}$. The gauge
condition also guarantees a simple expression for the metric. Namely if $\delta \mathcal X,\delta
\mathcal Y\in T_{\mathcal T}\mathcal M_{= \mathbf r}$ then
\begin{equation}\label{eq:TT-metric}
    \inp{\delta \mathcal X}{\delta \mathcal Y} = \sum_{\alpha=1}^d \inp{\delta \mathcal X_{(\alpha)}}{\delta \mathcal Y_{(\alpha)}}.
\end{equation}

\paragraph{Retraction.} The decomposition~\eqref{eq:tangent vector as rank 2r TT} of a tangent vector is also important for computing the retraction $R:{T\mathcal M_{=\mathbf r}\to \mathcal M_{\leq\mathbf r}}$ by HOSVD. If $\mathcal Y\in T_{\mathcal T}\mathcal M_{=\mathbf r}\subset \mathcal M_{\leq2\mathbf r}$, then $\mathcal T + t \mathcal Y \in \mathcal M_{\leq2\mathbf r}$ as well, since
\begin{equation}\label{eq:TT of X + T}
    (\mathcal T+t\mathcal Y)[i_1,\dots,i_d] = \begin{pmatrix}t\delta Y_{(1)}[i_1] & \mathcal U_{(1)}[i_1]\end{pmatrix}
    \begin{pmatrix}
        \mathcal V_{(2)}[i_2]         & 0                     \\
        t\delta \mathcal Y_{(2)}[i_2] & \mathcal U_{(2)}[i_2]
    \end{pmatrix}\cdots
    \begin{pmatrix}
        \mathcal V_{({d})}[i_{d}] \\\mathcal U_{({d})}[i_{d}]+t\mathcal Y_{({d})}[i_{d}]
    \end{pmatrix}.
\end{equation}
To obtain a retraction, we can thus define
\begin{equation} \label{eq:retract}
    R\colon T\mathcal M_{=\mathbf r}\to \mathcal M_{\leq\mathbf r},\qquad R(\mathcal T,t\mathcal Y) = \mathrm{HOSVD}(2 \mathbf r,\mathbf r)(\mathcal T+t\mathcal Y),
\end{equation}
where the $\mathrm{HOSVD}$ map was explained in Section~\ref{sec:ortho and hosvd}. Since the TT representation~\eqref{eq:TT of X + T} is explicitly available at no cost, the retraction $R$ can be computed in $O(d \, \max n_\alpha \, \max r_\alpha^3)$ flops; see~\eqref{eq:hosvd-cost} with the substitutions $r_\alpha \leadsto 2r_\alpha $ and $r'_\alpha \leadsto r_\alpha $.

\paragraph{Tangent space projection.} 

Given a tensor $\mathcal Z\in \rr^{n_1}\tensor\cdots\tensor \rr^{n_d}$, we want to compute its orthogonal projection $\mathcal P_{\mathcal T}(\mathcal Z)$ onto $T_{\mathcal T}\mathcal M_{= \mathbf r}$ directly in the representation~\eqref{eq:tangent-form}. This operation is needed for transporting a tangent vector to a different tangent space, and for calculating the Riemannian gradient of a loss function. In the first case, $\mathcal{Z}$ is a tensor of TT-rank at most $2\mathbf{r}$ as was noted above, while in the second, it is a sparse tensor since Problem~\ref{prob:tensorcompletion} is a tensor completion problem.

We first explain how to compute $\mathcal P_{\mathcal T}(\mathcal Z)$ efficiently when $\mathcal{Z}$ has TT-rank $\mathbf{r}' \leq 2\mathbf{r}$ and (non-normalized) cores $(\mathcal Z_{(1)},\dots,\mathcal Z_{(d)})$. Recall that $(\mathcal
U_{(1)},\dots,\mathcal U_{(d)})$ and $(\mathcal V_{(1)},\dots,\mathcal V_{(d)})$
are the left- and right-orthogonal cores of $\mathcal{T}$. 
First, we contract $\mathcal{Z}$ and $\mathcal{T}$ without its $\alpha$th core:
\begin{equation}\hspace{-2em}
    \delta \widetilde{\mathcal{Y}}_{(\alpha)} =
    \begin{tikzcd}[remember picture, column sep=3em, row sep=4ex, cells={nodes={draw=black,asymmetrical rectangle}}]
        \mathcal U_{(1)} \ar[r, dash, "r_1"] \ar[d, dash, "n_1"] &
        |[draw=white]|\cdots \ar[r, dash, "r_{\alpha-2}"]  &
        \mathcal U_{(\alpha-1)} \ar[r, dash, "r_{\alpha-1}"] \ar[d, dash, "n_{\alpha-1}"] &
        |[draw=black,dashed]|\hspace{2em}\vphantom{\mathcal Z_{(\alpha)}} \ar[r, dash, "r_{\alpha}"] \ar[d, dash, "n_{\alpha}"] &
        \mathcal V_{(\alpha+1)} \ar[d, dash, "n_{\alpha+1}"] \ar[r, dash, "r_{\alpha+1}"] &
        |[draw=white]|\cdots\ar[r,dash,"r_{d-1}"] &
        \mathcal V_{(d)} \ar[d, dash, "n_d"]
        \\
        \mathcal Z_{(1)} \ar[r, dash, "r'_1"] &
        |[draw=white]|\cdots \ar[r, dash, "r'_{\alpha-2}"] &
        \mathcal Z_{(\alpha-1)} \ar[r, dash, "r'_{\alpha-1}"] &
        \mathcal Z_{(\alpha)} \ar[r, dash, "r'_{\alpha}"] &
        \mathcal Z_{(\alpha+1)}  \ar[r, dash, "r'_{\alpha+1}"] &
        |[draw=white]|\cdots \ar[r, dash, "r'_{d-1}"] &
        \mathcal Z_{(d)}
    \end{tikzcd}\label{eq:tt-gradient-projection}
\end{equation}
The result of contracting this network is a tensor of shape $(r_{\alpha-1},n_\alpha,r_{\alpha})$. Next, we compute for $\alpha<d$
\begin{equation}  \label{eq:gauge-projection}
    \delta \mathcal Y_{(\alpha)} = \delta  \widetilde{\mathcal Y}_{(\alpha)} - \mathcal U_{(\alpha)}^L(\mathcal U_{(\alpha)}^L)^\top\delta  \widetilde{\mathcal Y}_{(\alpha)}^L=:P_\alpha\,\delta  \widetilde{\mathcal Y}_{(\alpha)},
\end{equation}
so that $\delta \mathcal Y_{(\alpha)}$ respects the gauge condition $(\mathcal U_{(\alpha)}^L)^\top \delta \mathcal Y^L_{(\alpha)} = 0$. For $\alpha =d$, we can take $P_d=I$ above so that $\delta \mathcal Y_{(d)}=\delta  \widetilde{\mathcal Y}_{(d)}$. The tuple $(\delta \mathcal Y_{(1)},\dots,\delta \mathcal Y_{(d)} )$ now defines $\delta \mathcal
Y = \mathcal P_{\mathcal T}(\mathcal Z)$ using~\eqref{eq:tangent-form}.

For efficiency reasons, it is worthwhile to pre-compute all products
$\mathcal U_{\leq \alpha}^\top \mathcal Z_{\leq \alpha}$ in a left-to-right sweep, and all products
$\mathcal Z_{\geq \alpha}^\top \mathcal V_{\geq \alpha}$ in a right-to-left sweep. Then, we
obtain $\delta \mathcal Y_{(\alpha)}$ by contracting $\mathcal U_{\leq \alpha-1}^\top \mathcal
Z_{\leq \alpha-1}$, $\mathcal Z_{(\alpha)}$ and $\mathcal Z_{\geq \alpha+1}^\top \mathcal V_{\geq
\alpha+1}$. Using the assumption $\mathbf r'\leq 2 \mathbf r$, an easy calculation shows that the total cost in flops of the vector transport is bounded 
by
\begin{equation}\label{eq:vector-transport-cost}
   6n_1r_1^2+4n_dr_d^2+\sum_{\alpha=2}^{d-1}14n_\alpha r_{\alpha-1}r_\alpha^2+8n_\alpha r_{\alpha-1}^2r_\alpha.
\end{equation}

A different strategy is needed when $\mathcal{Z}$ is sparse, that is, $\mathcal Z[j_1,\dots,j_d]\neq 0$ only for 
$(j_1,\dots,j_d)\in \Omega \subset \{1,\dots,n_1\}\times\dots\times \{1,\dots,n_d\}$. Different from~\eqref{eq:tt-gradient-projection}, the (non-normalized) first-order variation is now obtained from the contraction with an unstructured $\mathcal Z$:
\begin{equation}
    \delta \widetilde{\mathcal Y}_{(\alpha)} =
    \begin{tikzcd}[remember picture, column sep=2.2em, row sep=3ex, cells={nodes={draw=black,asymmetrical rectangle}}]
        \mathcal U_{(1)} \ar[r, dash, "r_1"] \ar[d, dash, "n_1"] &
        |[draw=white]|\cdots \ar[r, dash, "r_{\alpha-2}"]  &
        \mathcal U_{(\alpha-1)} \ar[r, dash, "r_{\alpha-1}"] \ar[d, dash, "n_{\alpha-1}"] &
        |[draw=black,dashed]|\hspace{2em}\vphantom{\mathcal Z_{(\alpha)}} \ar[r, dash, "r_{\alpha}"] \ar[d, dash, "n_{\alpha}"] &
        \mathcal V_{(\alpha+1)} \ar[d, dash, "n_{\alpha+1}"] \ar[r, dash, "r_{\alpha+1}"] &
        |[draw=white]|\cdots\ar[r,dash,"r_{d-1}"] &
        \mathcal V_{(d)} \ar[d, dash, "n_d"]
        \\
        |[draw=white]| \phantom{\mathcal Z}&
        |[draw=white]| \phantom{\mathcal Z}&
        |[draw=white]| \phantom{\mathcal Z}&
        |[draw=white]| \mathcal Z  &
        |[draw=white]| \phantom{\mathcal Z}&
        |[draw=white]| \phantom{\mathcal Z}&
        |[draw=white]| \phantom{\mathcal Z}&
    \end{tikzcd}
    \tikz[remember picture, overlay]{
        \draw[black] ([xshift=-5,yshift=0]\tikzcdmatrixname-2-1.north west) rectangle ([xshift=2,yshift=0]\tikzcdmatrixname-2-7.south east);
    }\label{eq:dense-gradient-projection}
\end{equation}
Applying the gauge conditions is again done as in~\eqref{eq:gauge-projection}. We therefore only need to explain how to compute the slices $\delta\widetilde{\mathcal Y}_{(\alpha)}[i]$ for $1\leq i\leq n_\alpha$. Let
\begin{equation}\label{eq:multi-index set sparse projection}
    \Theta_{\alpha,i} = \{(j_1,\dots,j_d)\in \Omega\mid j_\alpha=i\}.
\end{equation}
By inspecting~\eqref{eq:dense-gradient-projection}, we then obtain
\begin{align}
    \delta \widetilde{\mathcal Y}_{(\alpha)}[i] & = \sum_{(j_1,\dots,j_d)\in \Theta_{\alpha,i}} \mathcal Z[j_1,\dots,j_d]\left( \mathcal U_{(1)}[j_1]\cdots\mathcal U_{(\alpha-1)}[j_{\alpha-1}] \right)^\top \left( \mathcal V_{(\alpha+1)}[j_{\alpha+1}]\cdots \mathcal V_{(d)}[j_{d}] \right)]. \label{eq:dense-gradient-mat}
\end{align}
Since $\mathcal U_{(1)}[j_1]$ is a row vector and $\mathcal U_{(\beta)}[j_\beta]$ is a matrix for $\beta>1$, the first product in brackets is a row
vector. Since the second product is also a row vector, we thus decompose $\delta \widetilde{\mathcal Y}_{(\alpha)}[i]$ as a sum of $|\Theta_{\alpha,i}|$ rank-one matrices. By abuse of notation, we write $\mathcal Z[\Theta_{\alpha,i}]$ for the vector with all values of $\mathcal Z[j_1,\dots,j_d]$ for $(j_1,\dots,j_d)\in \Theta_{\alpha,i}$. Similarly, we write $\mathcal U_{\leq \alpha-1}[\Theta_{\alpha,i}]$ for the matrix with columns given by $\mathcal U_{(1)}[j_1]\cdots\mathcal U_{(\alpha-1)}[j_{\alpha-1}]$ for $(j_1,\dots,j_d)\in \Theta_{\alpha,i}$. We denote the matrix $\mathcal V_{\geq \alpha+1}[\Theta_{\alpha,i}]$ similarly. Then~\eqref{eq:dense-gradient-mat} becomes
\begin{equation} \label{eq:dense-gradient-einsum}
    \delta \widetilde{\mathcal Y}_{(\alpha)}[i] =
    \begin{tikzcd}[remember picture, column sep=0.8em, row sep=1.5ex, cells={nodes={draw=black,asymmetrical rectangle}}]
        |[draw=white]|&|[draw=white]|&\ar[ll,dash, "r_{\alpha-1}"'] \mathcal U_{\leq \alpha-1}[\Theta_{\alpha,i}] \ar[rr,dash,"|\Theta{_{\alpha,i}}|"] &
        |[draw=white]|\bullet &
        \mathcal V_{\geq \alpha+1}[\Theta_{\alpha,i}]\ar[rr,dash, "r_\alpha"] &|[draw=white]|&|[draw=white]|\\
        &&& \mathcal Z[\Theta_{\alpha,i}]\ar[u,dash,end anchor={[yshift=5]}]
    \end{tikzcd}
\end{equation}

The matrices $\mathcal U_{\leq \alpha-1}[\Theta_{\alpha,i}]$ and $\mathcal V_{\geq
\alpha+1}[\Theta_{\alpha,i}]$ can be computed efficiently for each $\alpha$ in, respectively, a
left-to-right and right-to-left sweep at a cost of
$|\Omega|\sum_{\alpha=2}^{d-1}r_{\alpha-1}r_\alpha$ flops. The total cost of computing  $\delta \widetilde{\mathcal Y}$ using~\eqref{eq:dense-gradient-einsum} is therefore given by
\begin{equation}\label{eq:cost-sparse-grad-proj}
    |\Omega|(r_1+r_{d-1}) + 3|\Omega|\sum_{\alpha=2}^{d-1} r_{\alpha-1}r_\alpha.
\end{equation}
Applying the gauge
conditions~\eqref{eq:gauge-projection} costs another $2n_1r_1^2+2\sum_{\alpha=1}^{d-1} n_\alpha
r_\alpha^2r_{\alpha-1}$ flops, assuming that the left- and right-orthogonalizations $\mathcal U_{(\alpha)}$, $\mathcal V_{(\alpha)}$ are already available.

\begin{remark}
The method above of projecting a sparse  $\mathcal
Z$ is mathematically equivalent to the one introduced
in~\cite{steinlechnerRiemannianOptimizationHighDimensional2016}, but ours is significantly more
efficient in practice. Instead of computing many rank-one updates in an outer loop over $\Omega$ as
in~\eqref{eq:dense-gradient-mat}, we group these operations into matrix products and tensor
contractions as in~\eqref{eq:dense-gradient-einsum}. The latter is much faster on modern hardware.
\end{remark}

\section{TTML: TT-based ML estimator}\label{sec:ttt}

We now describe how to solve the completion Problem~\ref{prob:tensorcompletion} using TTs to obtain a competitive ML estimator that we call TTML. The training procedure for TTML is summarized below in Algorithm~\ref{alg:ttt}. After training, inference is performed by evaluating the function $f_{\mathcal T,\mathbf t}$ defined by~\eqref{eq:tensorfunc}.
 
\begin{algorithm}[htb]
    \SetKwComment{tcc}{\# }{}
    \caption{Training the TTML estimator}\label{alg:ttt}

    \SetAlgoLined\KwIn{Training dataset $X_t\subset \rr^d,\,y_t\subset \rr$, validation dataset $X_v \subset \rr^d,y_v\subset \rr$,\hspace{100ex} auxiliary ML model $h\colon \rr^d\to\rr$, TT-rank $\mathbf r$.}
    \KwOut{Rank-$\mathbf r$ TTML estimator $f_{\mathcal T,\mathbf t}$} \vspace{1.5ex}

    $\mathbf t \leftarrow$ discretization of $\rr^d$ by binning training data $X_t$\tcc*[r]{see \S \ref{sec:thresholds}}

    $\mathcal T_0\leftarrow$ rank-$\mathbf r$ TT computed with TT-cross to approximate $h$ on the discretization $\mathbf t$\tcc*[r]{see \S\ref{sec:tt-cross}}

    \Repeat{Validation tensor-completion loss increases}{
        $\mathcal T_k\leftarrow$ update $\mathcal T_{k-1}$ using RCGD to minimize training tensor-completion loss\tcc*[r]{see \S\ref{sec:rcgd}}
    }
\end{algorithm}

\subsection{TT completion by RCGD}\label{sec:rcgd}
For the minimization of the loss function in Problem~\ref{prob:tensorcompletion}, we will use Riemannian conjugate gradient descent (RCGD). For a standard regression task, the loss function satisfies
\begin{equation} \label{eq:tt-tensor-completion}
    g(\mathcal T)=\sum_{i=1}^N \left(\mathcal T[j_1(x_i),\dots,j_d(x_i)] - y_i\right)^2.
\end{equation}
Here $X=(x_1,\dots,x_N)$, $y=(y_1,\dots,y_N)$ denotes the training data, and $j_\alpha(x_i)$ is defined by~\eqref{eq:tensorfunc}. Let us consider the index set 
\begin{equation} 
    \Omega = \{(j_1(x_i),\dots,j_d(x_i))\mid x_i\in X\},\label{eq:omega-equation}
\end{equation}
which also defines multi-index sets $\Theta_{\alpha,i}$ as used in~\eqref{eq:multi-index set sparse projection}.

To optimize~\eqref{eq:tt-tensor-completion} with RCGD, we compute the derivative of $g$. It is a sparse tensor $\nabla g$ with non-zero entries indexed by $\Omega$ and
\begin{equation}
    \nabla g[j_1(x_i),\dots,j_d(x_i)] = \mathcal T[j_1(x_i),\dots,j_d(x_i)]- y_i.\label{eq:tt-tc-grad}
\end{equation}
If we want to minimize the cross-entropy loss~\eqref{eq:tensorcompletion-crossentropy}, then we have instead
\begin{equation}
    \nabla g[j_1(x_i),\dots,j_d(x_i)] = \sigma(\mathcal T[j_1(x_i),\dots,j_d(x_i)])- y_i,\label{eq:tt-tc-grad-ce}
\end{equation}
with $\sigma$ the sigmoid function. Note that there may be multiple data points corresponding to the
same set of indices $(j_1,\dots,j_d)$. If this happens we simply sum over these data points in~\eqref{eq:tt-tc-grad} and~\eqref{eq:tt-tc-grad-ce}.

We summarize the general procedure of RCGD in
Algorithm~\ref{alg:tensor-completion-tt}. 

\begin{algorithm}[htb]
    \SetKwComment{tcc}{\# }{}
    \caption{RCGD for TT completion}\label{alg:tensor-completion-tt}

    \SetAlgoLined\KwIn{Training data $X,y$, thresholds $\mathbf t$, initial guess for TT $\mathcal T_0$}\vspace{1.5ex}

    $\xi_0 \leftarrow \mathcal P_{\mathcal T}(\nabla g)$\tcc*[r]{ $\mathcal P_{\mathcal T}$ projection of sparse Euclidean gradient, and $\nabla g$ as in~\eqref{eq:tt-tc-grad}}
    $\eta_0\leftarrow -\xi_0$\\
    $t_0\leftarrow$ result of line search along $\eta_0$\\
    $\mathcal T_1 \leftarrow R(\mathcal T_0, t_0\eta_0)$\\
    \For{$k=1,2,\dots$}{
        $\xi_k \leftarrow \mathcal P_{\mathcal T}(\nabla g)$\tcc*[r]{Gradient at step $k$}
        $\beta_k\leftarrow $ conjugate gradient scalar \\
        $\eta_k \leftarrow -\xi_k+\beta_k \mathcal P_{\mathcal T_k}(\eta_{k-1})$ \tcc*[r]{Search direction at step $k$}
        $t_k\leftarrow$ result of line search along $\eta_k$\tcc*[r]{Step size at step $k$}
        $\mathcal T_{k+1} \leftarrow R(\mathcal T_k, t_k\eta_k)$\tcc*[r]{Retract as in~\eqref{eq:retract}}
    }
\end{algorithm}

As line search, we use standard Armijo backtracking\footnote{We also experimented with non-monotone Armijo backtracking as well as backtracking using the Wolfe conditions, but this gave no improvement in performance for our application.} applied to $\phi_k(t)=g(R(\mathcal T_k, t\eta_k))$. Even in the context of Riemannian optimization, this can be implemented fairly easily; see, e.g., \cite{absilOptimizationAlgorithmsMatrix2008}. In particular, thanks the rigidity of $R$, we have $\phi_k'(0)=\inp{\xi_k}{\eta_k}$. 
In addition, choosing a good initial step
size $t_{k,0}$ for backtracking algorithm leads to a significant difference in
performance. In~\cite{steinlechnerRiemannianOptimizationHighDimensional2016}, the initial step is
computed based on an exact minimizer of the line search objective without retraction,
\begin{equation}
    t_{k,0,\text{lin}} = \argmin_t g(\mathcal T_k+t\eta_k) = \frac{\inp{\xi_k}{\eta_k}}{\inp{\eta_k}{\eta_k}}.
\end{equation}

We found that for our setting a better initial step is the (type-I) Barzilai-Borwein (BB) step
size. Let $s_{k} = t_{k-1}\mathcal P_{\mathcal T_k}(\xi_{k-1})$ and $y_k=\xi_k- \mathcal
P_{\mathcal T_k}(\xi_{k-1})$. Then, as in~\cite{wenFeasibleMethodOptimization2013}, we define
\begin{equation}
    t_{k,0,\text{BB1}} = \frac{\inp{s_k}{s_k}}{|\inp{s_k}{y_k}|}.
\end{equation}
This does however require computing the vector transport of the gradient, which is
unnecessary in $t_{k,0,\text{lin}}$.

For the conjugate gradient scalar $\beta_k$, we either use steepest descent $\beta_k=0$, or the
Riemannian Fletcher-Reeves choice\footnote{Both choices lead to similar performance.}
\begin{equation}
    \beta_{k,\text{FR}} = \frac{\inp{\xi_k}{\xi_k}}{\inp{\xi_{k-1}}{\xi_{k-1}}}.
\end{equation}

\begin{remark}\label{rem:als-dmrg-problems}
    The more standard ALS and DMRG approaches for solving this problem do not work well in this
    context. In the ALS scheme we optimize the TT one core at a time. The ALS objective for
    each core $\mathcal C^{(\alpha)}$ decouples into problems for each slice of the form
    \begin{equation}\label{eq:ALS-one-core}
    \mathcal C^{(\alpha)}[i] = \argmin_X \sum_k \left(\mathcal U_{\leq \alpha-1}[\Theta_{\alpha,i}[k]]^\top
    X \mathcal V_{\geq \alpha+1}[\Theta_{\alpha,i}[k]]-y[\Theta_{\alpha,i}[k]]\right)^2
    \end{equation}
    where, by abuse of notation, $y[\Theta_{\alpha,i}[k]]$ is the $k$th number $y_\ell$ such that
    $j_\alpha(x_\ell)=i$. This is a least-squares problem in the entries of $X$, but unfortunately
    it is often ill-conditioned. Observe that the slice $\mathcal C^{(\alpha)}[i]$ is of size
    $r_{\alpha-1}r_\alpha$. If thus $|\Theta_{\alpha,i}|<r_{\alpha-1}r_\alpha$, the problem is
    underdetermined. Additionally, the indices $j_\alpha(x_k)$ are not uniformly distributed, and it
    often happens that multiple data points $x_k$ end up in the same grid-cell. 
    This further reduces the effective number of data points available for determining the slice
    $\mathcal C^{(\alpha)}[i]$. In practice this means ALS can only be used if the number of thresholds
    is small for each feature, and if the ranks $r_\alpha$ are small. In our experiments, regularization techniques
    somewhat helped, but the lack of data remained a large issue. For DMRG, the situation is
    even worse, since the optimization problem for the supercores decouples even further and there are even less effective data points available per supercore. Gradient based algorithms do not suffer from
    this issue, and for this reason we have chosen to solve the tensor completion problem using Riemannian
    conjugate gradient descent instead.
\end{remark}

\subsection{Initialization from existing estimators with TT-cross}\label{sec:tt-cross}

The Riemannian algorithm described in Section~\ref{sec:rcgd} is only
effective with a good initialization. Unless the number of thresholds and TT-rank is small, the RCGD algorithm tends to end up in `bad' local minima with random initialization. We therefore propose to generate an initial guess by fitting a TT to an existing ML estimator $h\colon
\mathbb R^d\to \mathbb R$ trained on the data. Good estimators are, for example, random forests or neural networks. Fitting a TT to an estimator $h$ is
different from the tensor completion Problem~\ref{prob:tensorcompletion} because we can evaluate $h$ at any point we wish. The goal is therefore to solve\footnote{In practice, the norm
in~\eqref{eq:tensor-fit} is approximated as a sum over a
finite amount of points.}
\begin{equation}\label{eq:tensor-fit}
    \min_{\mathcal T\in \mathcal M} \|f_{\mathcal T,\mathbf t} -h\|^2.
\end{equation}
We will do this using the TT-cross approximation algorithm
\cite{oseledetsTTcrossApproximationMultidimensional2010, savostyanovFastAdaptiveInterpolation2011}.
There are two versions of the TT-cross algorithm, based on ALS or on DMRG that optimize one or two cores, respectively, of the TT at a time. A full sweep of the ALS (DMRG) version costs $O(dnr^3)$ ($O(dn^3r^3)$) flops and $O(dnr^2)$ ($O(dn^2r^2)$) evaluations of the function $h$, respectively. While the DMRG version is more expensive, it requires significantly fewer iterations to converge, can adapt the ranks, and tends to result in a better approximation of the function $h$. We recommend using the DMRG version unless it is prohibitively expensive. In all the experiments described in Section~\ref{sec:experiments} we have used the DMRG version.

A similar idea to improve the initialization of tensor completion problems was proposed in~\cite{kapushevTensorCompletionGaussian2020} where TT-cross was used in combination with a Gaussian process (GP) trained on
the same data. In our setting, we found that it is worth exploring different ML estimators for this initialization step since the final accuracy obtained depends strongly on the combination of the training data and the ML estimator. This
will be discussed further in Section~\ref{sec:experiments}.

\subsection{Overview of computational cost}\label{sec:complexity}

We give an overview of the computational costs in flops of all the operations needed for the TTML estimator.
For simplicity we assume that the TT is of order $d$ with uniform TT-rank $r_\alpha=r$ and uniform dimensions $n_\alpha=n$. Furthermore, we assume $n>r$ so that $r^3= O(nr^2)$.
For RCGD, we denote the number of 
(Armijo) line search steps used by $k_{\textrm{LS}}$.
Finally, $N$ is the number of data points in the training set.

\begin{center}
\begin{tabular}{lll}
\toprule
Operation & Complexity (flops) & Eq.\\
\midrule
Orthogonalization of TT & $2(d-2)nr^3+\OO(nr^2)$ & \eqref{eq:orth-complexity}\\
HOSVD from rank $r$ to $r'$ & $2r^3+2(d-2)(nr^2r'+nr(r')^2+2r^3)$ & \eqref{eq:hosvd-cost}\\
Inference (per sample) & $(d-2)r^2+\OO(d\log_2n + r)$ & \eqref{eq:TT-def}\\
Projection of sparse gradient & $3N(d-2)r^2+2(d-2)nr^3+\OO(nr^2)$ & \eqref{eq:cost-sparse-grad-proj}\\
Retraction & $12(d-2)nr^3+\OO(nr^2)$ & (\ref{eq:hosvd-cost}, \ref{eq:retract})\\
Vector transport & $22(d-2)nr^3+\OO(nr^2)$ & \eqref{eq:vector-transport-cost}\\ 
\midrule
RCGD (one step) & $(3+k_{\textrm{LS}})N(d-2)r^2+(24+12k_{\textrm{LS}})(d-2)nr^3$\\ 
 & $\qquad +\OO(k_{\textrm{LS}}nr^2+Nd\log_2n)$\\

ALS TT-cross (full sweep) & $\OO(dnr^3)$ flops and $\OO(dnr^2)$ function evaluations\\
DMRG TT-cross (full sweep) & $\OO(dn^3r^3)$ flops and $\OO(dn^2r^2)$ function evaluations\\
\bottomrule
\end{tabular}
\end{center}

Most importantly, observe that all of the operations are linear in $d$ and $N$. Since $N$ typically scales as $O(d)$ to avoid overfitting, we obtain a complexity of $O(d^2)$, like other tensor completion algorithms; see, e.g.,~\cite{steinlechnerRiemannianOptimizationHighDimensional2016}. We note that the $O(d\log_2n)$ term in prediction is due to computing the
indices $j_i(x_k)$ associated to data points. Since the thresholds are sorted, this is done using
binary search. The cost of one step of RCGD is computed as the sum of one sparse gradient
projection, $k_{\textrm{LS}}$ retractions and $k_{\textrm{LS}}\cdot N$ predictions, one vector transport, and one
orthogonalization. All other operations needed for RCGD are at most $O(k_{\textrm{LS}}nr^2)$.

\section{Experiments}\label{sec:experiments}

The source code and documentation for all the experiments described in this paper are available as a Python package at \href{https://github.com/RikVoorhaar/ttml}{https://github.com/RikVoorhaar/ttml}. All the experiments were performed on a desktop computer with an 8-core Intel i9-9900 using version 1.0 of the software. Our implementation of the TT-cross algorithm is a translation to Python of the MATLAB implementation in the TT-toolbox \cite{oseledetsTTtoolbox2014}.

\subsection{General setup and datasets used}\label{sec:general setup}

We have collected several datasets from the UCI Machine Learning
Repository~\cite{duaUCIMachineLearning2017} to evaluate our TTML estimator. All the datasets have a
relatively small number of features, and a modest number of samples. The used datasets are
summarized in Table~\ref{tab:datasets}.

\newcommand{\regr}{Regr.}
\newcommand{\class}{Class.}
\begin{table}[htb]
    \centerfloat
\begin{tabular}{llll}
    \toprule
                          Dataset & Task & \#Features & \#Samples \\
    \midrule
                        AI4I 2020 & \class &     6 &    10000  \\
               Airfoil Self-noise & \regr &      5 &     1503  \\
                   Bank Marketing & \class &     16 &    45211 \\
                            Adult & \regr &     14 &    32561  \\
    Concrete compressive strength & \regr &      8 &     1030   \\
           Default of credit card & \class &     23 &    30000  \\
    Diabetic Retinopathy Debrecen & \class &     19 &     1151  \\
        Electrical Grid Stability & \class &     13 &    10000  \\
            Gas Turbine Emissions & \regr &     10 &    36733   \\
                  Online Shoppers & \class &     17 &    12330  \\
       Combined Cycle Power Plant & \regr &      4 &     9568   \\
                    Seismic-bumps & \class &     15 &     2584  \\
                    Shill Bidding & \class &      9 &     6321  \\
                     Wine quality & \regr &     11 &     4898   \\
    \bottomrule
\end{tabular}
\caption{Table of datasets used in numerical experiments}\label{tab:datasets}
\end{table}

As baseline estimators, we chose random forest and boosted trees, since they both consistently perform very well on all the datasets. Implementations were provided by the Python packages \texttt{scikit-learn}~\cite{pedregosaScikitlearnMachineLearning2011} and \texttt{XGBoost} \cite{chenXGBoostScalableTree2016}, respectively. 

For all experiments, we created a random 70/15/15 train/validation/test split, where the validation data is used for early stopping and hyperparameter optimization. For each estimator and dataset, we first optimized the most relevant hyperparameters using the validation split\footnote{Using Bayesian optimization as in \cite{bergstraMakingScienceModel2013}} and kept only the best result. 
Using these best hyperparameters, the estimators were fitted and evaluated again on the same train/validation/test splits, and the error on the test set was recorded.  Due to the relatively low number of samples
in our datasets, this resulted in high variance. We therefore repeated this experiment 12 times on different train/validation/test splits, and we report the mean and standard deviation of the loss over these 12 splits. All estimators are evaluated on the same splits of the data, and the same procedure is used for both the baseline and our TTML estimators. 

For the TTML estimator, the main hyperparameters are the TT-ranks, number of thresholds and
estimator used for initialization. We used a uniform maximum number of thresholds for each feature,
and a uniform TT-rank for each core. The thresholds were chosen to divide the data values $X$ in
roughly equally sized bins along each axis, since this generally gives the best performance (as will
be discussed in Section~\ref{sec:thresholds}). We trained four different TTML estimators that
were each initialized with TT-cross (with fixed rank) using the following estimators\footnote{We
also tried Gaussian Processes as in~\cite{kapushevTensorCompletionGaussian2020} but the results were
not competitive for our datasets.}: random forests, multi-layer perceptrons (shallow neural
networks) with one or two layers(s), and boosted trees. This was then followed by several steps of
RCGD to minimize the loss as explained in Problem~\ref{prob:tensorcompletion}. During RCGD we
monitored performance on the validation dataset for early stopping.

The results on all datasets are shown in Table~\ref{tab:main-results}. 
From the table it is clear that the performance of TTML depends strongly on which estimator was used during the
TT-cross initialization. With the best initialization, TTML only improves on XGBoost once, but it does match or beat the random forest for 3 datasets. Overall, TTML does not seem to improve the baseline estimators if one is only interested in test error. However, as we will investigate below, TTML is superior in error versus model complexity and inference speed. 

Some of the datasets\footnote{The Concrete compressive strength, Wine quality, and Combined cycle power plant datasets, where they reported scores of respectively 29.8, 0.49, and 14.1.} were also used by~\cite{kargasSupervisedLearningEnsemble2020, kargasNonlinearSystemIdentification2020}. These authors similarly discretized ML estimators using low-rank tensors, but they use CPs instead of TTs. It is therefore interesting to compare the estimators. While TTML outperforms their reported scores on these three datasets, a direct comparison is not possible unfortunately since we do not have access to an implementation of their estimator.

\begin{table}[htb!]
\centerfloat
\begin{tabular}{lrrrr|rr}
    \toprule
    Dataset &    \texttt{ttml\_xgb} & \texttt{ttml\_rf} & \texttt{ttml\_mlp1} & \texttt{ttml\_mlp2} & \texttt{xgb} & \texttt{rf}\\
    \midrule
    AI4I 2020 &   0.23(15) &  \textbf{0.220(68)} & 0.238(98) &  0.35(15) & 0.0426(78) & 0.0490(76) \\
    Airfoil Self-noise &   2.63(72) &   2.81(62) &  \textbf{2.28(53)} &  2.39(70) &   2.39(67) &   3.45(56) \\
        Bank Marketing & \textbf{0.1911(65)} & 0.1938(58) & 0.255(43) &  0.84(62) & 0.1717(43) & 0.1753(46) \\
                 Adult &  0.369(24) &  \textbf{0.357(12)} & 0.396(28) &  0.90(45) & 0.2788(76) & 0.2984(75) \\
Concrete compr.~strength &  23.5(6.3) &  \textbf{23.4(7.5)} & 46.9(8.6) &    37(13) &  17.9(4.0) &  24.4(4.9) \\
Default of credit card &   \textbf{0.59(22)} &   0.59(29) &  0.83(20) &  1.04(28) & 0.4302(55) & 0.4275(60) \\
Diab.~Retinopathy Debr. &   3.08(55) &   \textbf{1.9(1.5)} &  5.7(4.1) &  6.0(1.6) &  0.573(47) &  0.578(31) \\
Electrical Grid Stability &  \textbf{0.029(23)} &  0.080(64) & 0.041(23) & 0.065(33) & 0.0011(29) & 0.0171(15) \\
 Gas Turbine Emissions &  \textbf{0.856(59)} &  1.028(86) &  2.67(29) &  3.88(89) &  0.291(12) &  0.434(26) \\
       Online Shoppers &   0.79(38) &   \textbf{0.58(20)} &  0.60(28) &  0.83(25) &  0.222(11) &  0.227(11) \\
Comb.~Cycle Power Plant &  \textbf{11.4(1.4)} &  11.6(1.3) & 14.6(5.3) & 12.5(1.5) &   8.7(1.1) &  11.2(1.2) \\
         Seismic-bumps &   0.85(11) &   1.12(54) &  1.89(51) &  \textbf{0.75(31)} &  0.228(28) &  0.219(28) \\
         Shill Bidding &  \textbf{0.041(31)} &  0.195(63) & 0.097(69) & 0.081(49) & 0.0075(50) & 0.0213(36) \\
          Wine quality &  0.510(20) &  \textbf{0.474(19)} &  1.53(49) &  3.4(2.1) &  0.388(13) &  0.380(13) \\
                        \bottomrule
\end{tabular}
\caption{Comparison of test error of TTML to XGBoost (\texttt{xgb}) and random forest (\texttt{rf}).
We performed TT-cross initialization using four different estimators: XGBoost (\texttt{ttml\_xgb}), random forest
(\texttt{ttml\_rf}), MLP with one and two hidden layer(s) (\texttt{ttml\_mlp1} and
\texttt{ttml\_mlp2}). The mean over 12 runs (with standard deviation in brackets) is shown for the
best choice of hyperparameters. For each dataset we mark the best initialization in bold.}
\label{tab:main-results}
\end{table}

\subsection{Test error, model complexity, and inference speed}\label{sec:speed}

We now investigate the relation between model complexity, inference speed, and test error in more detail for the \textit{airfoil self-noise} dataset.

Model complexity for the estimators is defined as the number of parameters in the model. For TTML, we used the number of entries in all the cores of the TT, which is $O(dr^2n)$ with $n$ the number of thresholds used per feature and $r$ the TT-rank. For a random forest and a boosted tree model, the number of parameters is proportional to the number of trees and the number of nodes in the tree. For both models, the decision trees are encoded as an array, and the number of parameters is computed as the sum of the sizes of these arrays for each tree in the random/boosted forest. The number of parameters of a multi-layer perceptron is computed as the total number of elements used for all the weights and biases. 

We trained the TTML estimator with TT-ranks ranging
from 2 to 19, and maximum number of thresholds per feature ranging from 15 to 150. For each pair of
rank and number of thresholds, we initialized as explained in the previous section with TT-cross. The best model in terms of test error was kept (this initialization has no effect on the number of
parameters or the inference speed). Furthermore, we trained a random forest, boosted tree, and MLP model
for a large range of hyperparameters on the same data. 

\begin{figure}[htb]
    \centerfloat
    \includegraphics[width=1.1\textwidth]{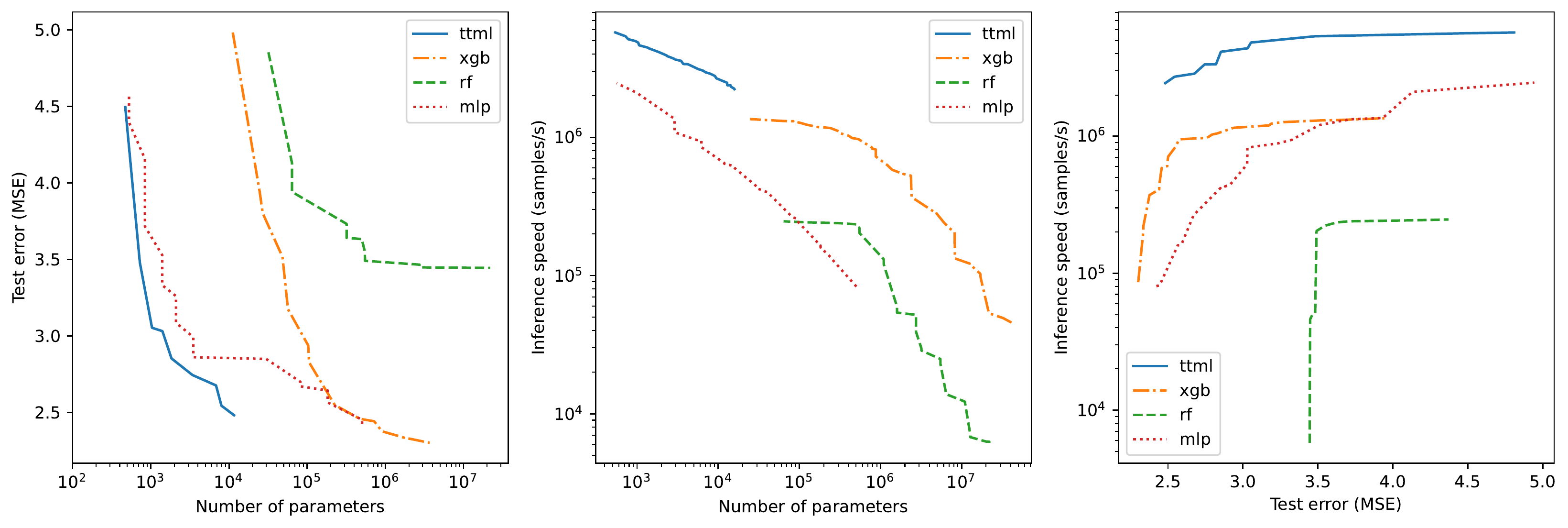}
    \caption{Relation between test error, model complexity and inference speed for different models:
    TTML (\texttt{ttml}), XGBoost (\texttt{xgb}), random forest
    (\texttt{rf}) and multilayer perceptron (\texttt{mlp}).}
    \label{fig:complexity}
\end{figure}

The results of these comparisons are visible in Figure~\ref{fig:complexity}. We can see that for a given test error our TTML model consistently performs faster, and needs
less parameters. This suggests that our estimator is potentially useful in low-memory applications
where inference speed of a pretrained model is favored over accuracy.

\begin{figure}[htb]
    \centering
    \includegraphics[width=1.0\textwidth]{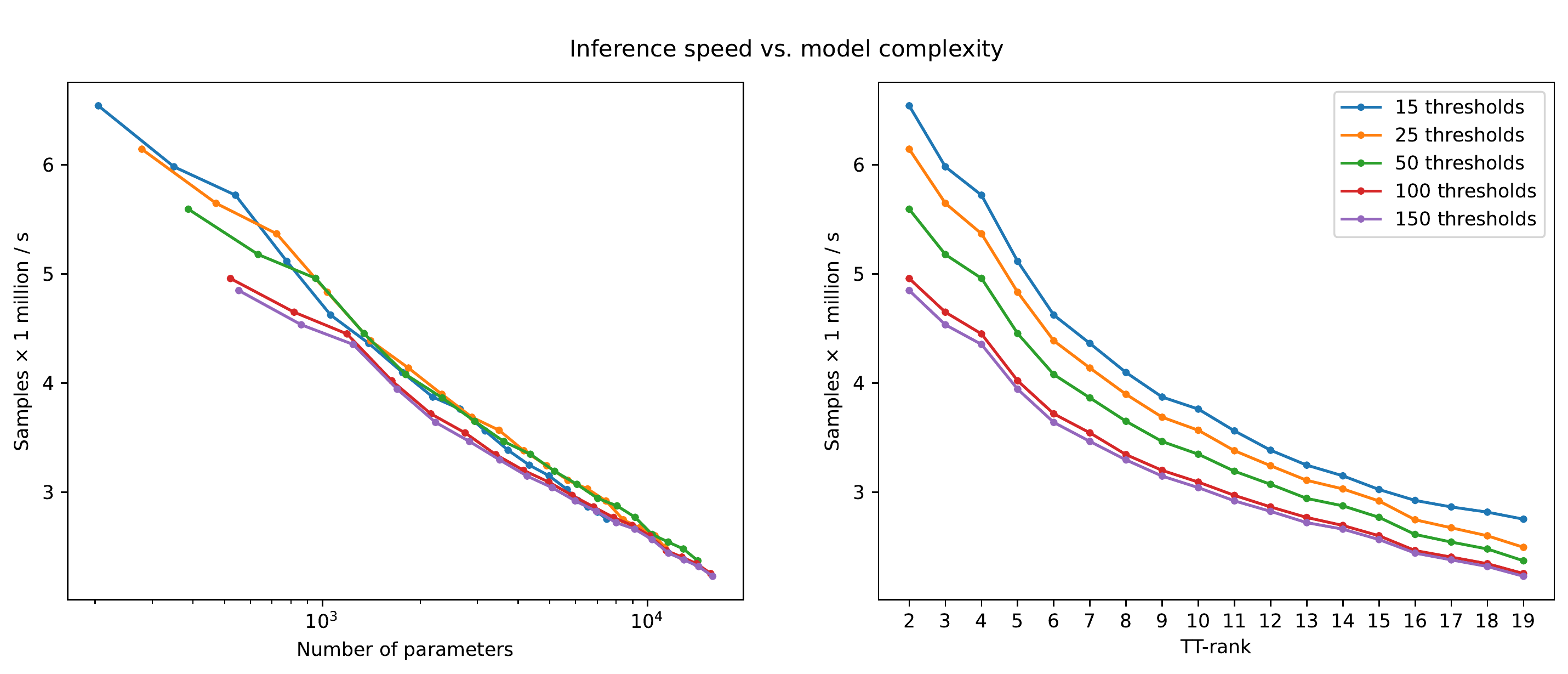}
    \caption{Inference speed in function of
    the number of parameters (left panel) and  the TT-rank (right panel) of the estimator.
    For each value of the TT-rank, we used 5
    different numbers of thresholds ranging from 15 to 150 (indicated with different colors).} 
    \label{fig:speed}
\end{figure}

In Figure~\ref{fig:speed} we show the relation between number of parameters, TT-rank and inference
speed in more detail. In theory, inference speed scales as $(d-1)r^2 + O(d\log_2 n+r)$; see Section~\ref{sec:complexity}. This  influence on $r$ is reflected in our experimental
results. The dependence on $n$ is only logarithmic since a binary
search is used to associate a correct multi-index to each data point. In practice, however, we observe a larger
dependence on $n$ than theoretically expected. This is likely because, in our implementation, inference is bound by memory speed, and memory usage depends linearly on $n$.

\subsection{Feature space discretization methods}\label{sec:thresholds}

As explained in Section~\ref{sec:tensor completion as ML estimator}, the feature space is discretized using the thresholds $\mathbf t$ so that a discrete tensor can be used as an ML estimator. Choosing a good feature space discretization turns out to be important for the quality of the estimator. The dependence of the estimator on the discretization is not
differentiable, and we only have heuristic methods for choosing the thresholds. 

A straightforward
way to choose the thresholds is to pick them such that they divide the data values $X$ for each
feature into equally sized bins in terms of frequency. That is, we choose them according to quantile
statistics of the data. Another discretization relies on choosing the thresholds using k-means to
solve a one-dimensional clustering problem for each feature; see
\cite{doughertySupervisedUnsupervisedDiscretization1995}. Instead of dividing the feature space directly, one can also first fit a random forest or decision tree to the
data, and then use the decision boundaries of the decision trees as thresholds for the
discretization. 

\begin{figure}[htb]
    \centering
    \includegraphics[width=0.65\textwidth]{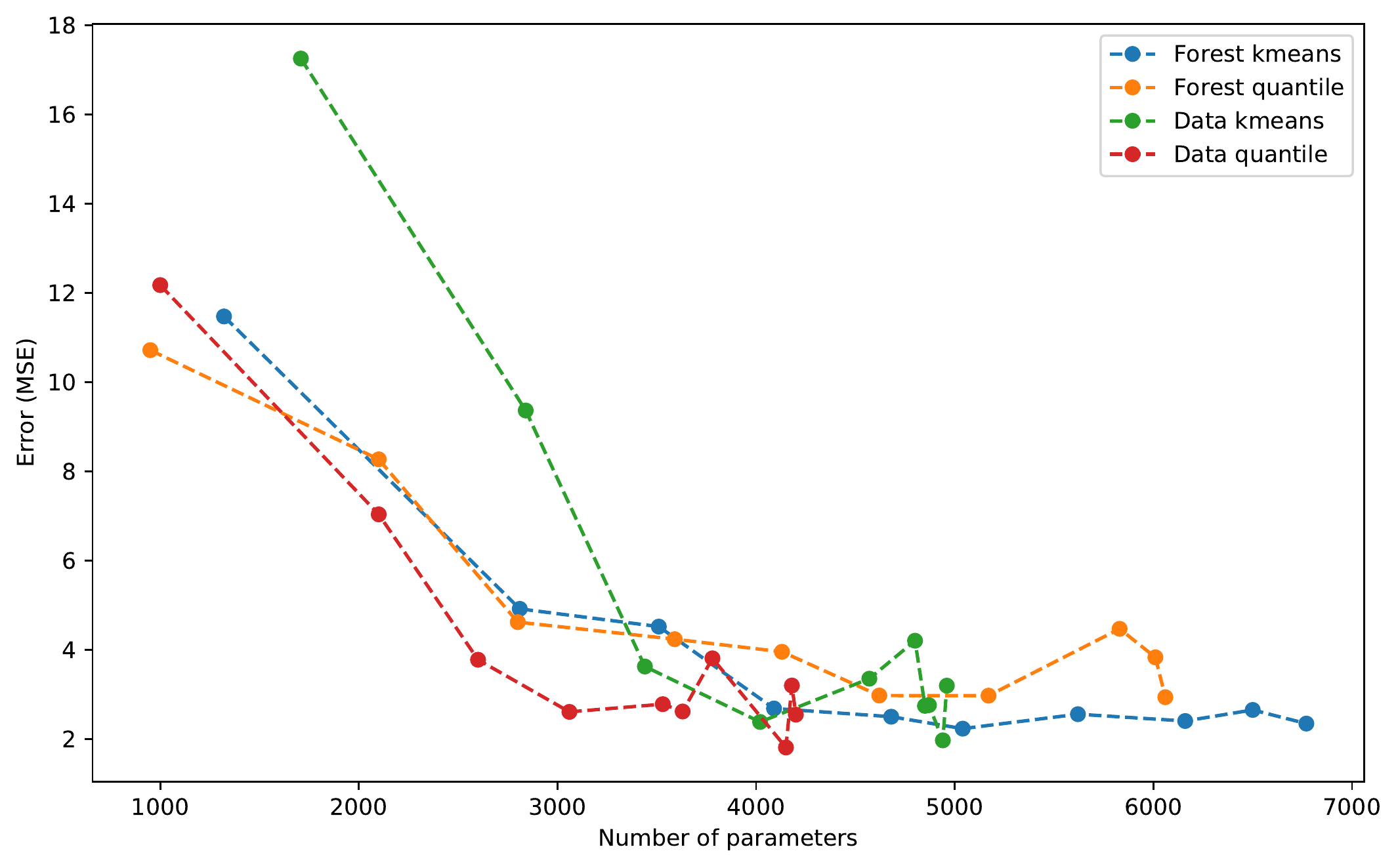}
    \caption{Comparison of different feature space discretization methods. We trained a TTML with TT-rank 10 using a random forest for initialization  on the airfoil self-noise dataset. We trained the estimator 15 times for each discretization method, and for each target number of thresholds per feature. We then recorded the median performance on the test error for each run of 15.}
    \label{fig:discretization}
\end{figure}

In Figure~\ref{fig:discretization}, we compared these four discretization methods for the \textit{airfoil self-noise} dataset. In particular, we trained a random forest on the data, and then used either quantile binning or
k-means clustering to select a subset of thresholds, either directly from the features or from the decision boundaries. For this data set, we see that the simplest methods, directly quantile binning of the features, works best. Since similar behavior was observed for other data sets as well, we choose this discretization in all our tests.

\subsection{Ordering of the features}

Some estimators, like neural networks and random forests, are insensitive to permutation of
the features. This is not the case for TTML since the TT-rank of a tensor, and thus our model, depends on this ordering. In some applications with tensors, certain orderings (or more general, certain topologies of the tensor network) lead to much better approximations given a budget of parameters; see, e.g, \cite{ballaniTreeAdaptiveApproximation2014, bebendorfSeparationVariablesFunction2014, grelierLearningTreebasedTensor2019}. In this section, we show experimentally that the same phenomenon can happen for TTML. However, in our setting, we will illustrate this for the test error and not for the approximation error of the underlying tensor.

For this experiment we again used the \textit{airfoil self-noise} dataset. Since it has only 5
features, we can train TTML estimators for all 120 different permutations of the features. We repeated the experiment on 30 different train/test/validation splits. For each permutation and
split, we trained our estimator with 40 thresholds per feature and a TT-rank of 6. The tensor
train was initialized on a random forest fit on the same training split. For each permutation the
mean performance is determined.\footnote{The variance in the mean performance $x$ is modeled as $\sigma(x)=\sigma_0 x$ with $\sigma_0=0.0294$. This value of $\sigma_0$ was obtained by dividing the mean sample standard deviation over all 120 permuations by the mean performance over all 120 permuations. We found that the experimental data is very consistent with this model, unlike computing sample standard deviation for each permutation separately or using a constant $\sigma(x)=\sigma_0$.}

\begin{figure}[htb]
    \centering
    \includegraphics[width=0.6\textwidth]{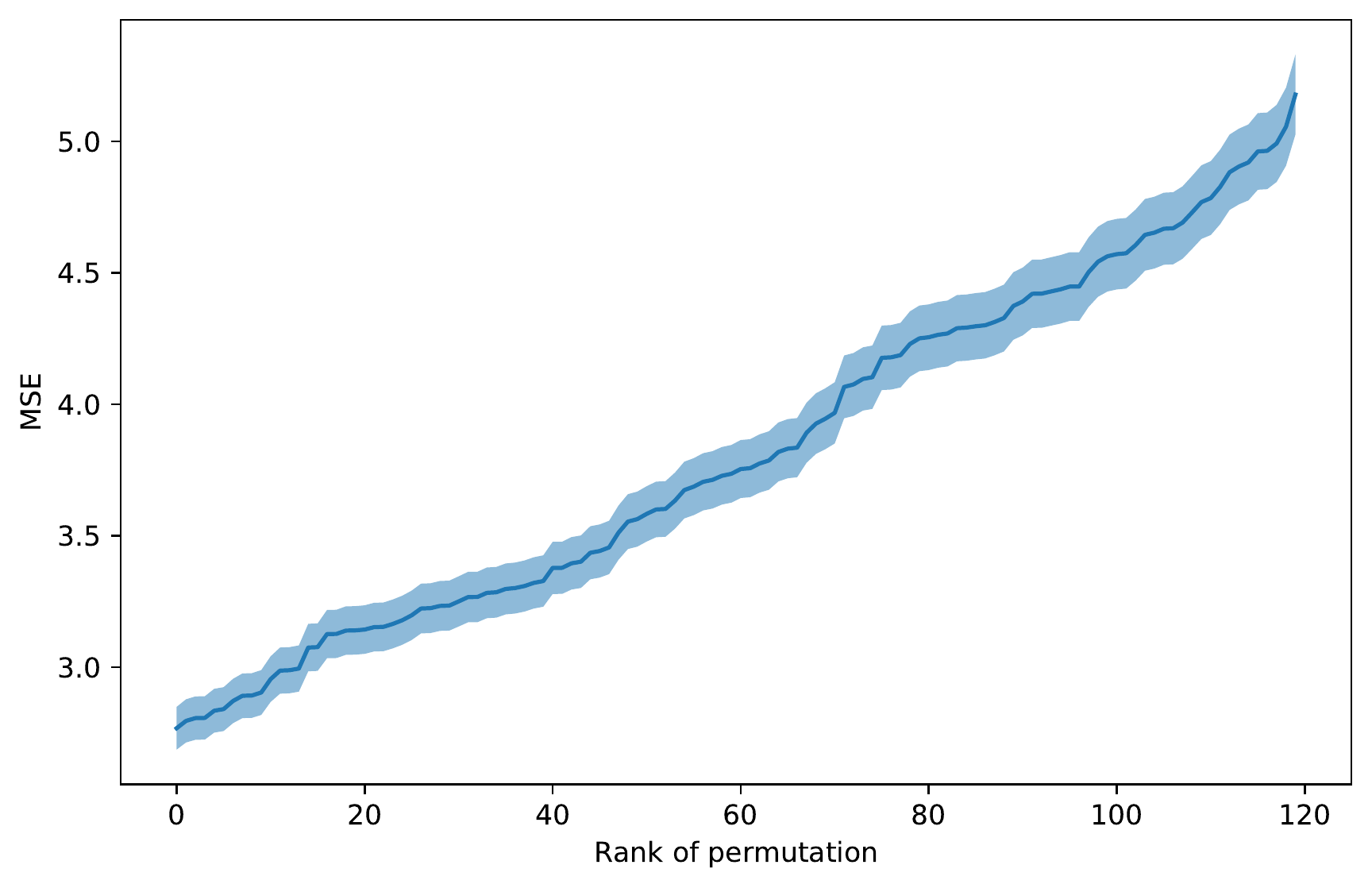}
    \caption{Sorted mean performance our estimator on all 120 permutations of the features on the
    \texttt{airfoil} dataset. The shaded area indicates the uncertainty in the mean.}
    \label{fig:permutation}
\end{figure}

The results are shown in Figure~\ref{fig:permutation}. We clearly see that the
ordering has a strong effect on the test error. The best 5 permutations are $(2, 1, 3, 4, 0), (0, 4, 2, 1, 3), (3,
2, 1, 4, 0), (0, 4, 1, 3, 2), (1, 3, 2, 4, 0)$, whereas the 5 worst are $(1, 0, 3, 2, 4), (4, 1, 2,
0, 3), (4, 2, 3, 0, 1), (3, 0, 2, 1, 4), (3, 0, 2, 4, 1)$. The only statistically significant
pattern we found is that feature zero is best put at position 0 or 4, especially compared to putting
this feature in the middle. Despite a non-optimal choice of hyperparameters, the performance of the best permutation beats the best performance on the `default' permutation of the features shown in Table~\ref{tab:main-results}. It is therefore clear that a method for estimating the optimal permutation of the
features would be very useful.\footnote{Such methods exists for \emph{compression} of TTs (see, e.g.,~\cite{ballaniTreeAdaptiveApproximation2014, bebendorfSeparationVariablesFunction2014, grelierLearningTreebasedTensor2019}) but it is not clear how these ideas can be extended when one is interested in test error.}

\subsection{Number of thresholds and TT-rank}\label{sec:thresholds-rank}

As last experiment, we investigate in more detail the dependence of the TTML estimator on the number of thresholds, the TT-rank, and the initialization method on the following datasets: \textit{airfoil self noise}, \textit{concrete compressive strength}, and \textit{shill bidding}. The training of TTML and the baseline methods (random forest and boosted tree), the optimization of the hyperparameters and the selection of the final methods was done as in Section~\ref{sec:general setup} for the construction of Table~\ref{tab:main-results}.

\begin{figure}[p]
    \centering
    \includegraphics[width=\textwidth]{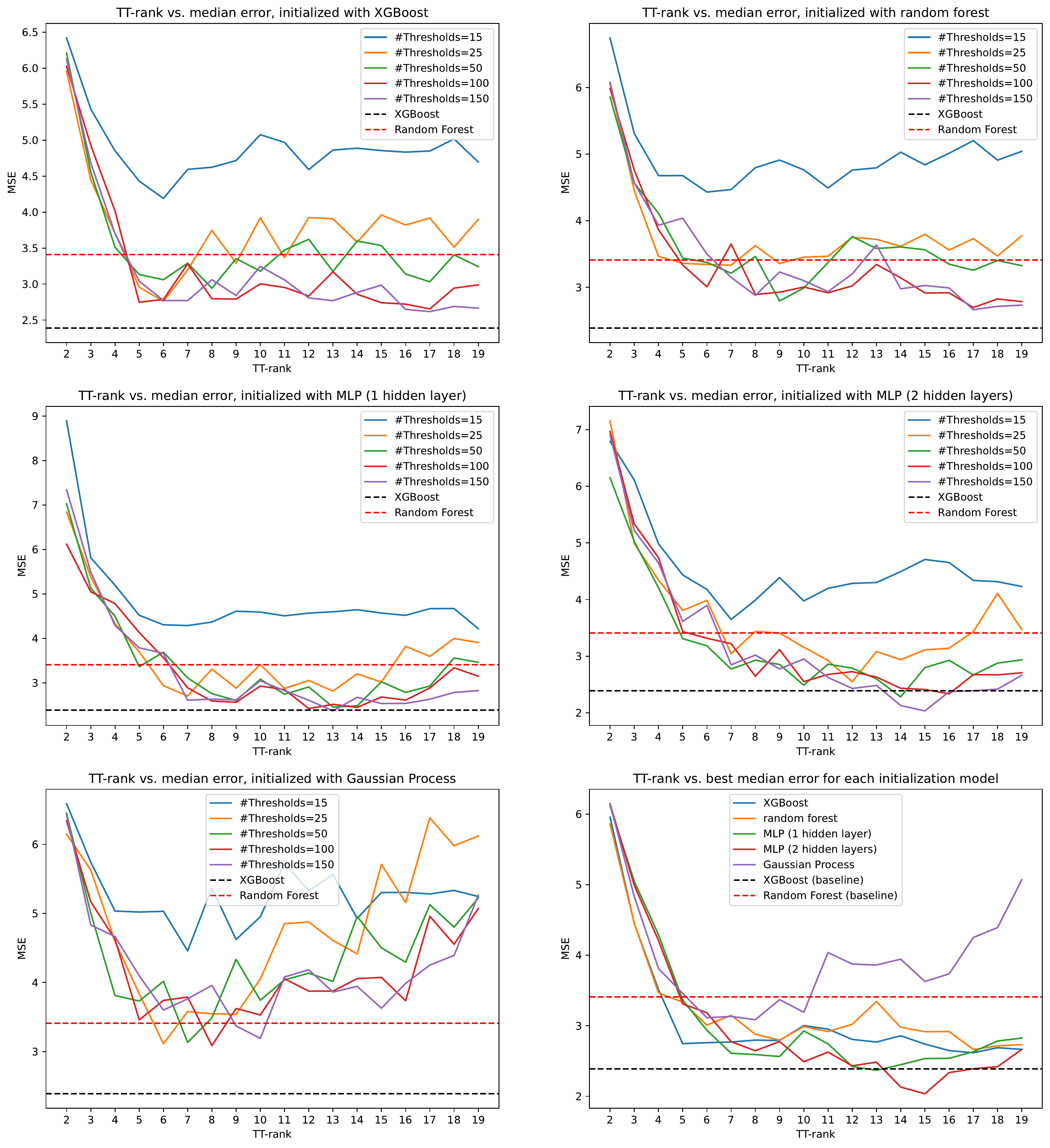}
    \caption{Comparison of TTML to XGBoost and random forest on the
    ``Airfoil Self-Noise'' dataset. Each panel has as different initialization for the TTML estimator and depicts several numbers of thresholds. \label{fig:airfoil-main}}
\end{figure}

\begin{figure}[p]
    \centering
    \includegraphics[width=\textwidth]{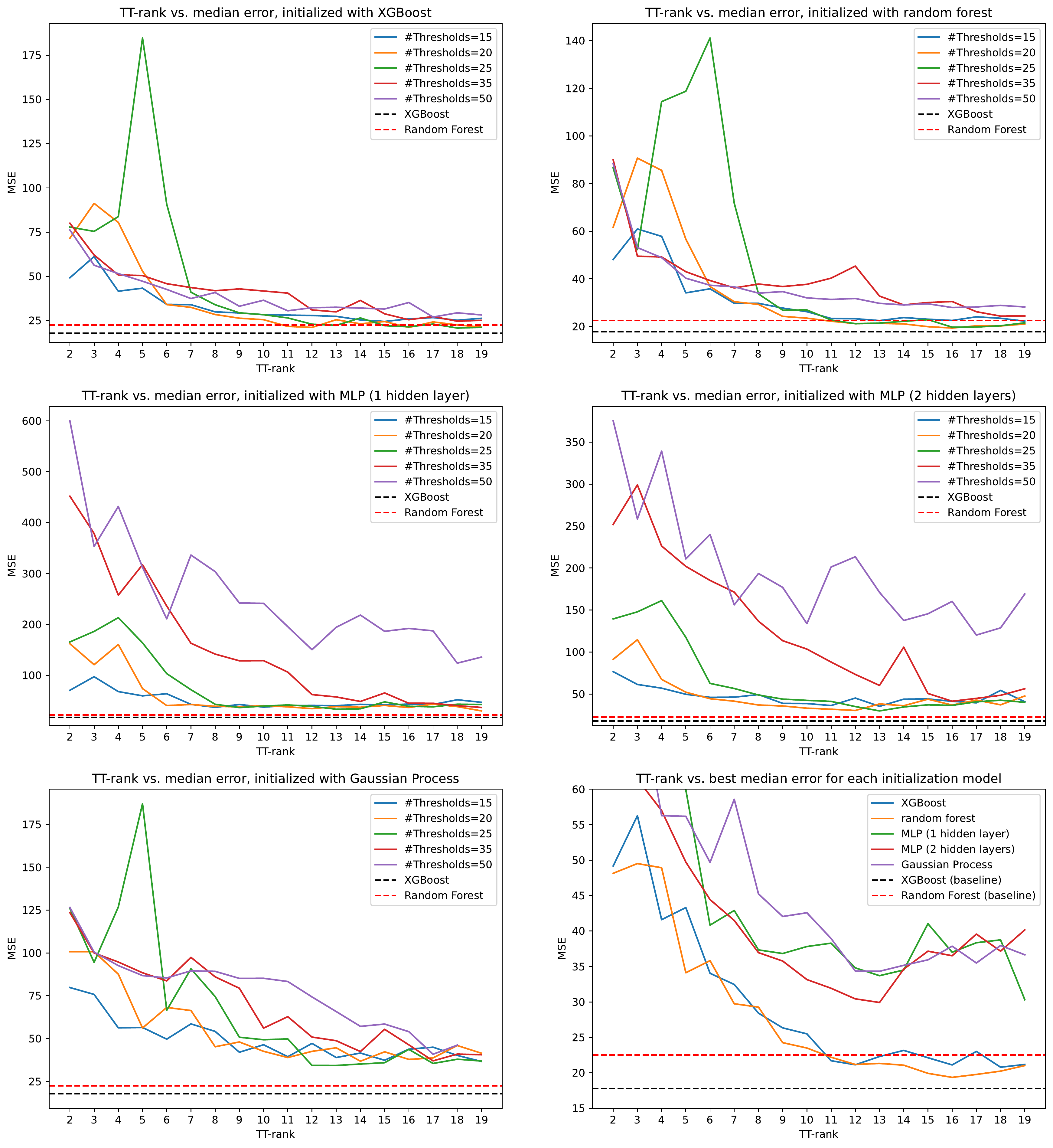}
    \caption{Same setting as in Figure~\ref{fig:airfoil-main} but for the
    ``Concrete Compressive Strength'' dataset. \label{fig:conrete-main}}
\end{figure}

\begin{figure}[p]
    \centering
    \includegraphics[width=\textwidth]{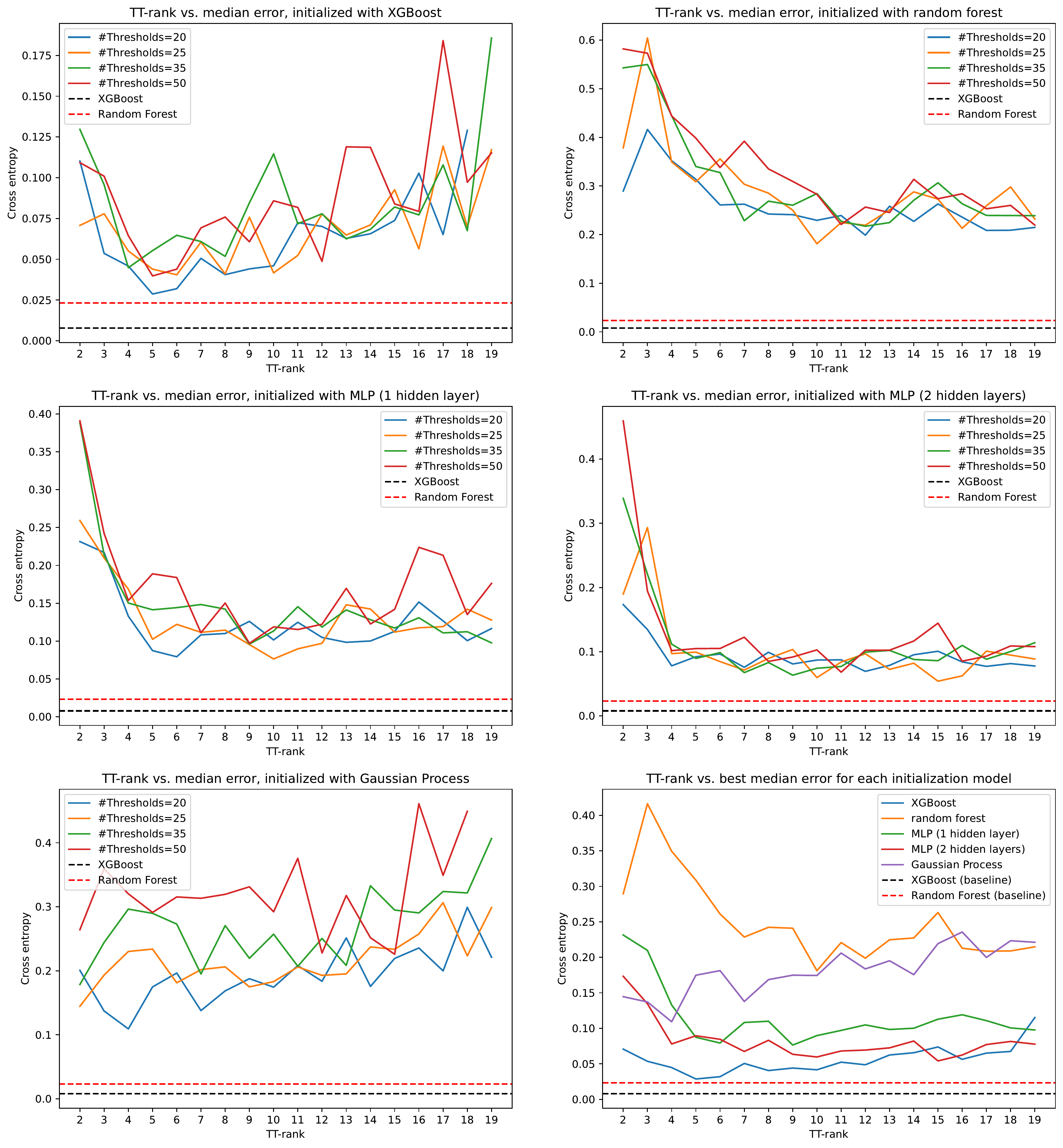}
    \caption{Same setting as in Figure~\ref{fig:airfoil-main} but for the
    ``Shill Bidding'' dataset. \label{fig:shill-bidding-main}}
\end{figure}

The results are visible for each dataset separately in Figures \ref{fig:airfoil-main}--\ref{fig:shill-bidding-main} further below. 
Most importantly, we see that the best initialization method is strongly dependent on
the dataset. For example, for the \textit{airfoil self noise} dataset, we obtain best performance
using an MLP for initialization, whereas for the \textit{concrete compressive strength} dataset it is achieved using a random forest, and for the \textit{shill bidding}
dataset it is XGBoost. Finally, Gaussian process based initialization is
consistently the worst. 

The performance is also dependent on the number of thresholds used for
initialization, and the TT-rank. The optimal number of thresholds and TT rank depends on the dataset and the estimator used for initialization. Generally the error decreases with TT-rank and number of thresholds until an optimal point after which the error (slowly) increases with additional complexity. It seems that the optimal TT-rank does not depend strongly on the number of thresholds used, and these hyperparameters can therefore effectively be optimized using a coordinate descent procedure.

\section{Extensions, future work, and conclusions}

\subsection{Extensions of TTML}

The TTML estimator described in Algorithm~\ref{alg:ttt} is very flexible, and can be generalized in many ways to create potentially more useful ML estimators. We will now discuss some of the more promising extensions, all of which may be studied in future work.

\paragraph{Topology of the network.}
Problem~\ref{prob:tensorcompletion} can be formulated for virtually any tensor format, not just TTs. An obvious generalization of the present work is therefore to replace TTs in Algorithm~\ref{alg:ttt} with a general \textit{tree tensor networks}, like the \textit{hierarchical Tucker} (HT) format that uses binary trees. Compared to TT, more general trees sometimes results in lower complexity (ranks) at the expense of learning a much more complicated topology of the tree; see, e.g.,~\cite{ballaniTreeAdaptiveApproximation2014, bebendorfSeparationVariablesFunction2014, grelierLearningTreebasedTensor2019}. In order to adapt our methods to such a different tensor format, we need a good black box approximation method, and an efficient implementation of retraction and vector transport to perform RCGD (or an alternative method to efficiently solve the tensor completion problem in the given format). For the HT format, black box approximation is described in~\cite{ballaniBlackBoxApproximation2013}, whereas (Riemannian) tensor completion for HT is described in~\cite{dasilvaOptimizationHierarchicalTucker2014, rauhutTensorCompletionHierarchical2015}. Adapting our methods to the HT format should therefore be straightforward. For other tensor formats, such as CP, tensor rings or MERA, black box approximation and tensor completion may be more challenging to perform with reasonable computational and storage complexity. 

\paragraph{Approximation of multivariate functions.}
While for simplicity we have only considered approximation of univariate functions $\rr^d\to\rr$, our methods readily generalize to the multivariate case $\rr^d\to \rr^m$. Given data $X=\{x_1,\dots,x_N\}\subset \rr^d$, $y=\{y_1,\dots,y_N\}\subset \rr^m$, we can use the same discretization $\mathbf t$ of $\rr^d$ as in the univariate case. We then consider the tensor completion problem for tensors $\mathcal T\in\mathcal M\subset \rr^{|\mathbf t_1|\times\cdots\times |\mathbf t_d|\times m}$. Using the same definition, the function $f_{\mathcal T, \mathbf t}$ defined by~\eqref{eq:tensorfunc} now takes values in $\rr^m$ and the regression tensor completion problem~\ref{prob:tensorcompletion} becomes

\begin{equation}
    \min_{\mathcal T\in \mathcal{M}}\sum_{i=1}^N \sum_{k=1}^m\left(\mathcal T[j_1(x_i),\dots,j_d(x_i),k] - y_i[k]\right)^2.
\end{equation}

There is no obstruction to using the TT format (or tree tensor networks) in this context; we simply add one additional core to the TT on the right. TT-cross can still be used to perform black box approximation, and RCGD can still be used for tensor completion in this context.

\paragraph{Use as first layer in feedforward network.} The multivariate generalization described previously can be used as first layer in a feedforward network using backpropagation in order to improve its performance. The function $f_{\mathcal T,\mathbf t}\colon \rr^d\to\rr^{m_1}$ can be composed with any other parametric model $h_\theta\colon \rr^{m_1}\to\rr^{m_2}$. Here $h_\theta$ can be any differentiable model, such as a linear model or a neural network. This results in the optimization problem
\begin{equation}\label{eq:composite-opt}
\min_{\mathcal T\in \mathcal M, \theta}\sum_{i=1}^N L(h_\theta\circ f_{\mathcal T,\mathbf t}(x_i),y_i),
\end{equation}
where $L$ is some loss function that depends on the training data $X,y$. This can be efficiently optimized using gradient based methods and backpropagation. We first compute the gradients of $h_\theta$:
\begin{equation}
g_i := \nabla_zh_\theta(z)\big|_{z=f_{\mathcal T,\mathbf t}(x_i)} \in \rr^{m_1}. 
\end{equation}
Then similar to~\eqref{eq:tt-tc-grad} we compute the Euclidean gradient for $L$ with respect to $\mathcal T$ by
\begin{equation}
    \nabla_{\mathcal T}L(h_\theta\circ f_{\mathcal T,\mathbf t}(x_i),y_i)=\sum_k \mathcal T[j_1(x_i),\dots,j_d(x_i),k]g_i[k].
\end{equation}
After projection onto the tangent space $T_{\mathcal T}\mathcal M$ this gradient can then be used for Riemannian gradient descent as usual. Note that since $f_{\mathcal T,\mathbf t}$ is a locally constant function, the derivative $\nabla_x f_{\mathcal T,\mathbf t}(x)=0$ a.e., and TTML therefore does not support backpropagation itself. For this reason, TTML can only be used as a \textit{first} layer in a feedforward network.

\subsection{Conclusions}

We have described a novel general-purpose ML estimator based on tensor completion of TTs. The estimator is competitive compared to existing estimators if measured by inference speed and model complexity, for a given test error. The model is very flexible, and has several promising generalizations. This illustrates the expressive power of low-rank tensor decompositions, and their potential use in data science and statistics. From another perspective, we have also highlighted the importance of good initialization for tensor completion problems, and we have shown that using auxiliary models for initialization can improve the performance on tensor completion problems dramatically.

\bibliography{ttml}
\bibliographystyle{apalike}

\end{document}